\documentclass[lettersize,journal]{IEEEtran}
\usepackage[colorlinks,urlcolor=blue,linkcolor=blue,citecolor=blue]{hyperref}
\usepackage{algorithm}
\usepackage{algorithmic}
\usepackage{verbatim}
\usepackage{soul}
\usepackage{url}
\usepackage[utf8]{inputenc}
\usepackage{amsmath}
\usepackage{amsthm}
\usepackage{booktabs}
\usepackage{amsfonts}
\usepackage{mathrsfs}
\usepackage{array}
\usepackage{multirow}
\usepackage{graphicx}
\usepackage{subcaption}
\ifCLASSOPTIONcompsoc
  \usepackage[nocompress]{cite}
\else
  \usepackage{cite}
\fi

\ifCLASSINFOpdf
\else
\fi

\begin{document}

\title{HGV4Risk: Hierarchical Global View-guided Sequence Representation Learning \\for Risk Prediction}
\author{Youru Li,
        Zhenfeng Zhu$^{*}$,
        Xiaobo Guo,
        Shaoshuai Li,
        Yuchen Yang \\
        and Yao Zhao,~\IEEEmembership{Sensor~Member,~IEEE}
\IEEEcompsocitemizethanks{
\IEEEcompsocthanksitem Youru Li, Zhenfeng Zhu and Yao Zhao are with the Institute of Information Science, Beijing Jiaotong University, Beijing, China and the Beijing Key Laboratory of Advanced Information Science and Network Technology, Beijing, China. Zhenfeng Zhu is the corresponding author. \protect\\
E-mail: $\{$liyouru, zhfzhu, yzhao$\}$@bjtu.edu.cn
\IEEEcompsocthanksitem Xiaobo Guo and Shaoshuai Li are with MYBank, Ant Group, Beijing \& Hangzhou, China. \protect\\
E-mail: $\{$jefflittileguo.gxb, lishaoshuai.lss$\}$@antgroup.com
\IEEEcompsocthanksitem Yuchen Yang is with the Department of Biology, Johns Hopkins University Zanvyl Krieger School of Arts and Sciences, Baltimore, MD, USA\protect\\
E-mail: yuchen.yang@jhu.edu
}
% \thanks{Manuscript received -- --, ----; revised -- --, ----.}
}

\markboth{}%
{Shell \MakeLowercase{\textit{et al.}}: A Sample Article Using IEEEtran.cls for IEEE Journals}


\maketitle

\begin{abstract}
Risk prediction, as a typical time series modeling problem, is usually achieved by learning trends in markers or historical behavior from sequence data, and has been widely applied in healthcare and finance. In recent years, deep learning models, especially Long Short-Term Memory neural networks (LSTMs), have led to superior performances in such sequence representation learning tasks. Despite that some attention or self-attention based models with time-aware or feature-aware enhanced strategies have achieved better performance compared with other temporal modeling methods, such improvement is limited due to a lack of guidance from global view. To address this issue, we propose a novel end-to-end \underline{H}ierarchical \underline{G}lobal \underline{V}iew-guided (HGV) sequence representation learning framework. Specifically, the Global Graph Embedding (GGE) module is proposed to learn sequential clip-aware representations from temporal correlation graph at instance level. Furthermore, following the way of key-query attention, the harmonic $\beta$-attention ($\beta$-Attn) is also developed for making a global trade-off between time-aware decay and observation significance at channel level adaptively. Moreover, the hierarchical representations at both instance level and channel level can be coordinated by the heterogeneous information aggregation under the guidance of global view. Experimental results on a benchmark dataset for healthcare risk prediction, and a real-world industrial scenario for Small and Mid-size Enterprises (SMEs) credit overdue risk prediction in MYBank, Ant Group, have illustrated that the proposed model can achieve competitive prediction performance compared with other known baselines.
\end{abstract}


\section{Introduction}
\IEEEPARstart{A}{iming} at predicting the probability of future events, such as whether a patient will develop a disease or not\cite{DBLP:journals/tcyb/LiLGLHY22}, the risk prediction modeling is commonly used in the world of medicine \cite{DBLP:journals/jms/FijackoCGKCCS21,DBLP:journals/tkde/CoronatoC22} to help guide clinical decision-making but are also used in other fields such as finance and insurance \cite{DBLP:conf/cikm/LiYSD20,DBLP:conf/ijcai/YeQX20,DBLP:journals/tcss/ZhangZ22}. As a typical time series modeling task \cite{DBLP:conf/kdd/LuoYXM20}, risk prediction models usually make decision by learning trends of marker \cite{DBLP:conf/aaai/MaGWZWRTGM20} or sequential behavior \cite{DBLP:conf/sdm/Wang0ZCFJF021} from historical data. Recent studies show that some deep learning techniques, particularly Recurrent Neural Networks (RNNs) and their variants \cite{DBLP:conf/nips/HochreiterS96,DBLP:conf/emnlp/ChoMGBBSB14}, have achieved impressive performance in such a sequence data representation learning task \cite{Harutyunyan2019}.

Beyond plain RNNs, some studies have started to utilize attention mechanism to characterize the importance of the individualization in some general tasks such as computer vision \cite{DBLP:conf/nips/MnihHGK14} and natural language processing \cite{DBLP:journals/corr/BahdanauCB14}. Inspired by this, others have employed attention-based models \cite{DBLP:conf/ijcai/QinSCCJC17} for structured data representation learning. Such typical time-aware models focus on learning weights for each time interval and capturing the long-term temporal dependencies at instance level. However, they fail to explore more granular information encoded in the original channel-level signal. To solve such problem, models in another branch such as RETAIN \cite{DBLP:conf/nips/ChoiBSKSS16} and RetainEX \cite{DBLP:journals/tvcg/KwonCKCKKSC19} are proposed by exploring both temporal relationships and variable significance with a two-level neural attention framework. 

In this context, however, the irregularity of time intervals in historical sequence has not been addressed. To this end, some works \cite{DBLP:conf/kdd/BaytasXZWJZ17} have been proposed to introduce decay modules for jointly learning time-decay and contextual dependencies during the representation learning with irregular time series data, achieving better prediction performance. Although aforementioned studies have addressed both long-term dependencies and time irregularity, the effect is still far from satisfactory. Self-attention architecture has been introduced in models such as SAnD \cite{DBLP:conf/aaai/SongRTS18} and HiTANet \cite{DBLP:conf/kdd/LuoYXM20} to further capture the sequential dependencies in time series data, thus achieving better prediction performance. To explore the relationship between dynamic information and static information, ConCare \cite{DBLP:conf/aaai/MaZWRWTMGG20} has further improved the performance by modeling both channel-wise sequential dependencies and feature-dependencies while utilizing statics information. 

\begin{figure}[t]
	\centering
	\includegraphics[width=0.45\textwidth,height=5.5cm]{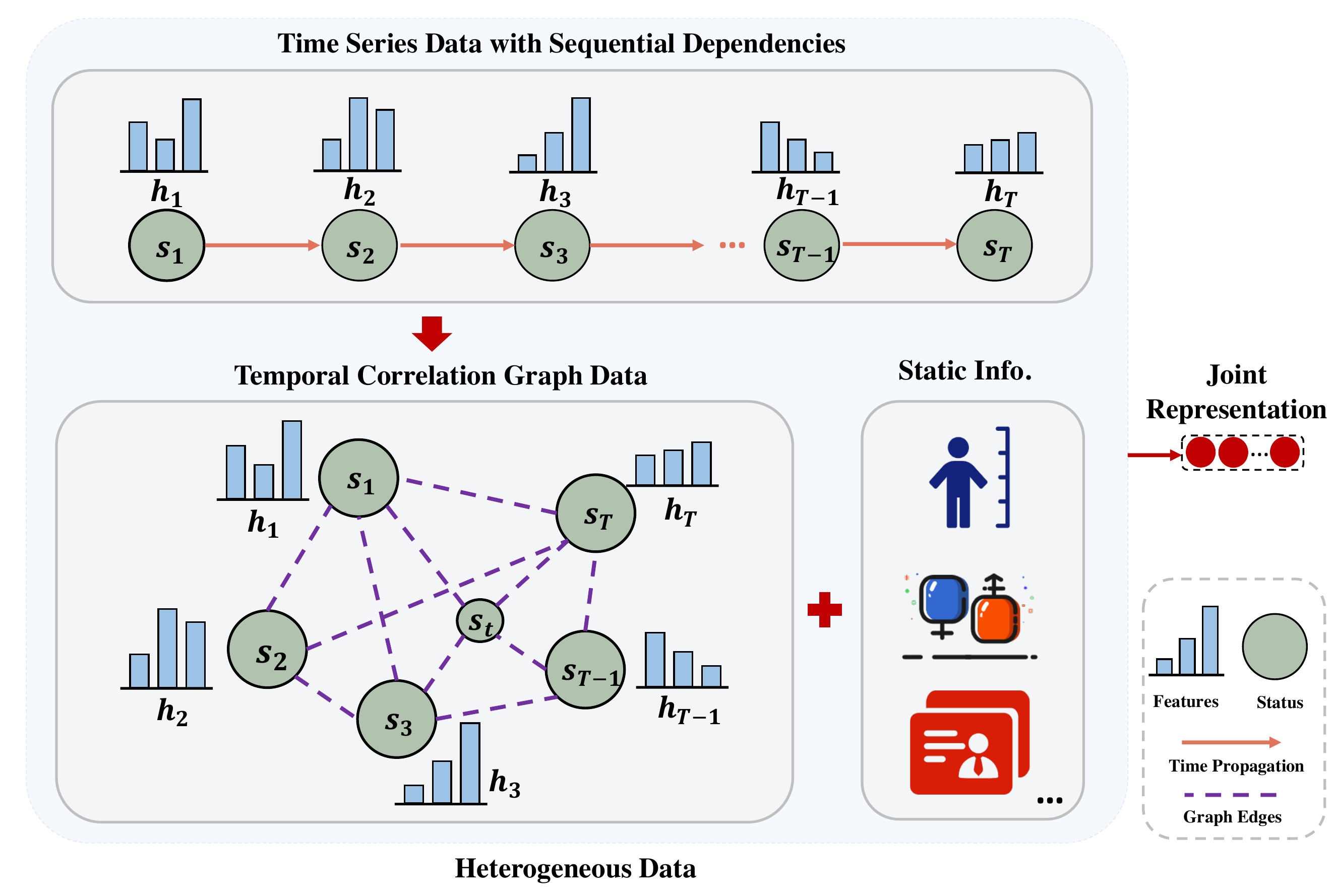}
	\caption{A demonstration of learning a joint representation from dynamic, static information and the temporal correlation graph induced from dynamic time series data. 
	}
	\label{fig1}
\end{figure}

Most of the above-mentioned temporal modeling work rely on progressive manners, which typically captures the sequential dependencies merely from the time series data. From this perspective, although dynamic and static information is available, some existing studies are yet to be done in capturing inherent patterns other than sequential dependencies in time series data over long time spans. This will to some extent help to reveal the temporal rhythmic variation of the observed status in time series, which can be characterized by the temporal correlation graph induced from dynamic time series data, as shown in Fig. \ref{fig1}. For these heterogeneous data, a representation learning framework should be established for jointly modeling both sequential dependencies in time series data and the global status correlation in temporal correlation graph data, meanwhile adopting static information.

In general, the major difference between our work and the above models is that we provide a new perspective for risk prediction modeling with the help of hierarchical representations learned from heterogeneous data. Given the limitations of existing methods, a workable model needs to meet three main challenges: 1) how to learn the temporal correlation among status in sequence data at instance level; 2) how to capture the specific patterns against long-term decay from channel-level irregular time series data in an explicit and end-to-end manner; 3) how to aggregate the channel-wise representations and instance-wise representations together with static information by a union hierarchical heterogeneous representation learning framework. Therefore, to address these issues, this paper makes the following contributions:
\begin{itemize}
	\item {For the purpose of obtaining sequential clip-aware representation from temporal correlation graph data at instance level, a global graph embedding method is proposed.}
	\item {A novel key-query attention, i.e., the  harmonic $\beta$-attention, is proposed to learn a global trade-off between time-aware decay and observation significance for irregular time series data at channel level adaptively.}
	\item{To coordinate the hierarchical representation learned from both instance-level and channel-level data, a heterogeneous information aggregation strategy is introduced for modeling dependencies among the multi granularity information.}
	\item{We advance the representation learning framework HGV by testing it on two real-world risk prediction tasks across both healthcare and financial domains, and show its effectiveness.}
\end{itemize}

\section{Related Work}
In recent years, risk prediction has been successfully applied in many real-world tasks, especially in healthcare and financial areas \cite{DBLP:journals/tcyb/ChoiCY17,DBLP:conf/kdd/0001ZWWFTWLWH21,DBLP:conf/ijcai/YangZZWSZFY020,zhang2022credit}. Indiscriminately, as a sequence representation learning problem, here, we only take the healthcare risk prediction as an example, and summarize the following typical modeling paradigms:

\textbf{Time-aware Models.}
As a typical time-aware model, RETAIN \cite{DBLP:conf/nips/ChoiBSKSS16} claims that existing deep learning methods often have to face the challenge of trade-off between interpretability and performance. It developed two RNNs in the reversed time and then generated the context vector with an attention-based representation learning module. Although it can improve interpretability to some extent, its performance is limited. Moreover, as a patient subtyping model, T-LSTM \cite{DBLP:conf/kdd/BaytasXZWJZ17} was proposed to address the challenge that the traditional LSTMs suffer from suboptimal performance when handling data with time irregularities. It learns a subspace decomposition of the cell memory, which enables time decay to discount the memory content according to the elapsed time. Although they have achieved better performance, it lacks of considering the impact of time-aware decay \cite{DBLP:conf/aaai/MaZWRWTMGG20}.

\textbf{Attention-based Models.}
Prior efforts usually leverage attention or self-attention architecture in risk prediction models. For example, as  a framework composed of an attention-based RNN and a conditional deep generative model, MCA-RNN \cite{DBLP:conf/icdm/LeePJM18} was proposed for capturing the heterogeneity of EHRs by considering essential context into the sequence modeling. To effectively handle long sequences in a time-series modeling task, SAnD \cite{DBLP:conf/aaai/SongRTS18} has employed a masked self-attention mechanism and introduced positional encoding and dense interpolation strategies for incorporating temporal order. Furthermore, for the better exploration of the personal characteristics during the sequences and the improvement of the time-decay assumption for covering all conditions, ConCare \cite{DBLP:conf/aaai/MaZWRWTMGG20} was proposed by combining a new time-aware attention and multi-head self-attention with a cross-head decorrelation loss. However, as most of the temporal modeling methods have done, they are also ineffective at capturing inherent patterns such as temporal correlations among status in time series data.

\textbf{Knowledge-enhanced Models.}
Due to the sparsity and low quality of data, some knowledge-enhanced models have been raised recently. The models such as GRAM \cite{DBLP:conf/kdd/ChoiBSSS17}, KAME \cite{DBLP:conf/cikm/MaYXCZG18}, MMORE \cite{DBLP:conf/ijcai/SongCYCFP19} and HAP \cite{DBLP:conf/kdd/ZhangKAC20} were proposed to improve the prediction performance on healthcare-related tasks by incorporating the medical ontology such as ICD codes. Moreover, to better model sequences of ICD codes, HiTANet \cite{DBLP:conf/kdd/LuoYXM20} assumes the non-stationary disease progression and proposes a hierarchical time-aware attention network to predict diagnosis codes by employing a time-aware Transformer at visit level and a time-aware key-query attention mechanism among timestamps. However, such knowledge encoded in medical ontology is not always available for all issues \cite{DBLP:conf/aaai/MaZWRWTMGG20}. Furthermore, to fully extract the correlation between similar patients inside the dataset, GRASP \cite{DBLP:conf/aaai/ZhangGMWWT21} clusters patients who have similar conditions and results, and then improves the performance by leveraging knowledge extracted from these similar patients. However, this risk prediction model leads to insufficient improvement from noisy knowledge due to the inevitable gap between the extracted knowledge and the patients themselves.

\begin{table}[t] 
    \renewcommand\arraystretch{1.3}
	\centering
	\caption{Notations and description.}   
	\begin{tabular}{p{1.5cm}<{\centering}|p{6.2cm}<{}} 
		\hline  
		Notation & Description \\
		\hline 
		$U=\{u_{i}\}$ & The set of $|U|$ users/patients \\  
		\hline
		$y_{i}, \tilde{y}_i$  & The ground truth and predicted labels for $u_{i}$ \\ 
		\hline
		$T$ & No. of time steps for making observation \\
		\hline
		$N_d$ & No. of channels for making observation \\
		\hline
		$S^{i}$ & The dynamic status information for $u_{i}$ \\  
		\hline
		$S^i_{\cdot t}\in\mathbb{R}^{N_d}$ & The observed statuses at the $t$-th time step \\
		\hline
		$S^i_{n \cdot}\in\mathbb{R}^{T}$ & The observed statuses from the $n$-th channel  \\ 
		\hline
		$f^{i}_{b}\in\mathbb{R}^{N_b}$  & $N_b$- dimensional basic features for $u_{i}$ \\  
		\hline
		$g_{i}\in\mathbb{R}^{T\times T}$ & Temporal clip global correlation graph using $\{S^{i}_{\cdot t}\}_{i=1,\cdots, T} $ as nodes\\
		\hline
		$E_{d}(\cdot),E_{b}(\cdot)$ and $E_{g}(\cdot)$ & Embeddings for $S^i_{n \cdot}$, $f^{i}_{b}$, and $g_{i}$, respectively\\
		\hline
		$\alpha(\cdot)$ & An attention weight vector \\
		\hline  
	\end{tabular} 
	\label{t1} 
\end{table} 

\section{Notations and Framework}
\newtheorem{myDef}{Global View}
\newtheorem{myDef1}{Channel View}
\newtheorem{MYdef}{Definition}

\subsection{Notations}
To facilitate the elaboration of the risk prediction task to be dealt with, some notations used throughout the paper are given first in Table \ref{t1}. 

Let $U=\{u_1,u_2,\ldots,u_{|U|}\}$ be the set of users/patients with the observed dynamic status information $S=\{S^{i}\in \mathbb{R}^{N_d \times T}\}_{i=1,\cdots,|U|}$ and the corresponding ground truth label\footnote{Here, the risk prediction is formulated as a binary classification problem, and without loss of generality, it is trivial to extend our model to multilevel risk prediction} $y=\{y_i\}\in \{0,1\}^{|U|}$, where $N_d$ and $T$ represent the number of channels and the time step for feature observation of time series, respectively, and $|U|$ denotes the volume of $U$. In addition, it is also assumed that a $N_b$-dimensional basic features (also called static information) $F_{b}=\{f^i_{b} \in \mathbb{R}^{N_b}\}_{i=1,\cdots,|U|}$ is also available for each user/patient together with the dynamic information $S$, e.g., the static information including age and weight, and so on, in healthcare risk prediction. Particularly, for the dynamic status information $S^i=\left[S^i_{n,t}\right]_{n,t} \in \mathbb{R}^{N_d\times T}$ of the $i$-th user/patient, we use the row vector $S^{i}_{n\cdot}=[s^{i}_{n,1}, s^{i}_{n,2},\ldots,s^{i}_{n,T}]\in \mathbb{R}^{T}$ of $S_i$ denote the observed $T$ historical sequence of statuses from the $n$-th channel, and likewise, the column vector $S^{i}_{\cdot t}=[s^{i}_{1,t}, s^{i}_{2,t},\ldots,s^{i}_{N_d,t}]^T\in \mathbb{R}^{N_d}$ of $S^i$ denote the observed historical statuses from $N_d$ channels at the $t$-th time step. With the given historical statuses $S^{i}$, a global status correlation graph $g_{i}$ can be built by using $\{S^i_{\cdot t}\}_{t=1,\cdots,T}$ as the nodes. With the above notations, we formally define the global view and channel view on characterizing a patient/user as follows.
\begin{MYdef}[Global View]
The global view at instance level is defined as a fused observation of both the global status correlation graph $g_{i}$ and static information $f^i_{b}$ from the $i$-th instance.
\end{MYdef}
\begin{MYdef}[Channel View]
The channel view is defined as a globally coordinated observation of each status $s^{i}_{n,1}, s^{i}_{n,2},\ldots,s^{i}_{n,T}$ in the historical sequence of statuses $S^{i}_{n\cdot}$ from the $n$-th channel.
\end{MYdef}
As opposed to the representations of the users/patients from global view, the channel view provides a more granular characterization of them.

\begin{figure*}[ht!]
	\centering
	\includegraphics[width=\linewidth,height=13cm]{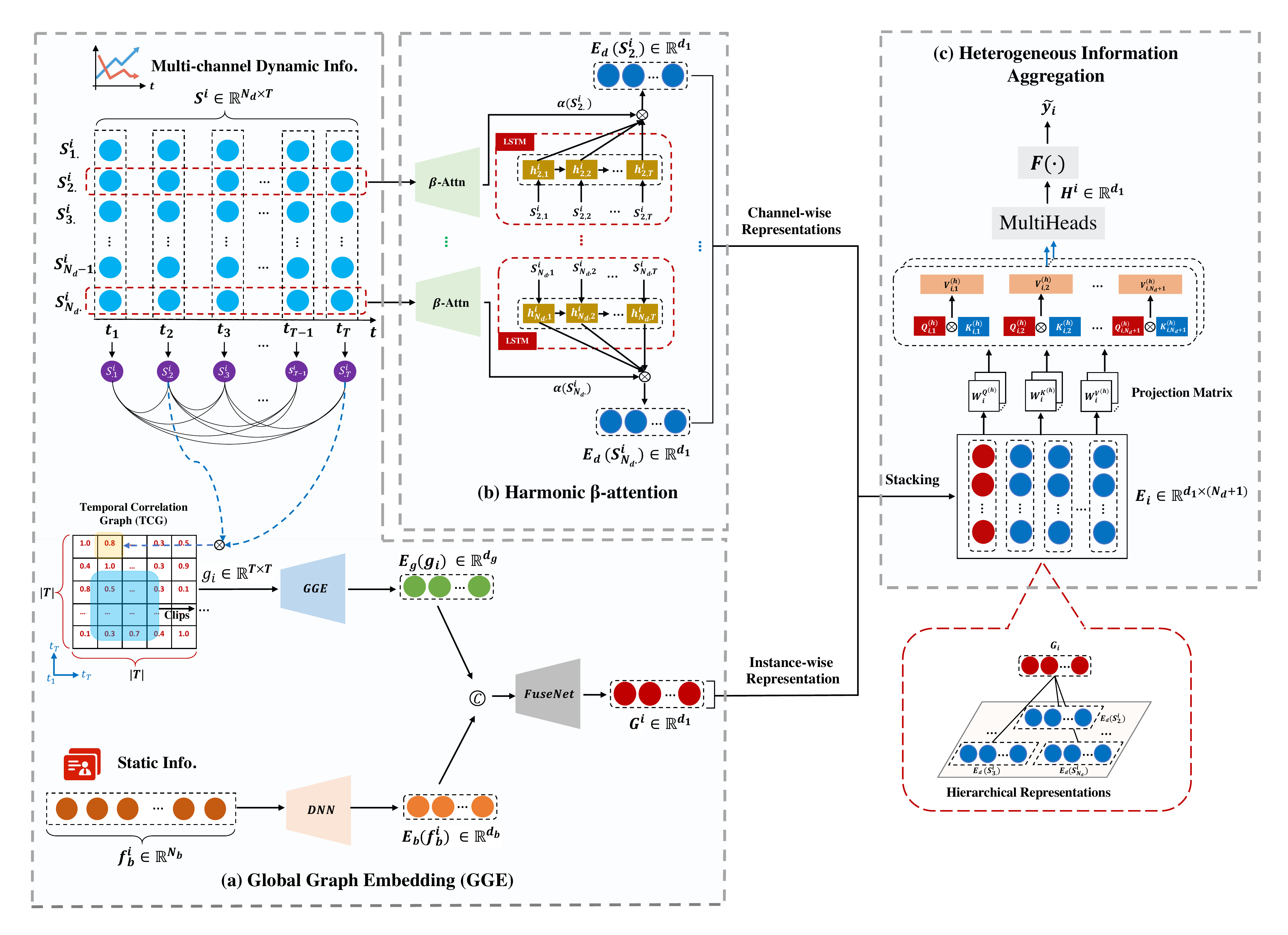}
	\caption{Graphical illustration of hierarchical global views-guided sequence representation learning for risk prediction. It is mainly composed of three modules: (a) \textbf{Global Graph Embedding (GGE)}; (b) \textbf{Harmonic $\beta$-attention}; (c) \textbf{Heterogeneous Information Aggregation}. These modules are trained as a whole in an end-to-end manner.}  
	\label{fig2}
\end{figure*}

\subsection{Overall Framework} 
A graphical illustration of the proposed hierarchical global view-guided sequence representation learning for risk prediction is given in Fig. \ref{fig2}. Specifically, the proposed framework mainly consists of three modules: 
\begin{enumerate}
    \item {\textbf{Global Graph Embedding (GGE)}}. It aims to learn global clip-aware representations via convolution neural network on a temporal correlation graph at the sequential clip level;
    \item{\textbf{Harmonic $\beta$-attention}}. Inspired by the F-score, the proposed harmonic $\beta$-attention attempts to make a global trade-off between time-aware decay and observation significance;
    \item{\textbf{Heterogeneous Information Aggregation}}. To aggregate heterogeneous information, the instance-wise representations and channel-wise representations are weighted and formed a unified representation with multi granularity information, which makes the hierarchical guidance on two global views are well coordinated. 
\end{enumerate}

\section{Methodology}
\subsection{\textbf{Global Graph Embedding}}
Distinguished from the sequential dependencies in time series data, the correlations between temporal status should also be considered from a global perspective, thus to enhance the acquisition of more valuable historical statuses for the risk prediction in the future. For this purpose, a global graph embedding approach is proposed.

\subsubsection{\textbf{Temporal Correlation Graph}}
Empirically, in medical diagnosis, the physical signs of the human body at different times can be similar and interrelated to each other. For an example, a person's blood glucose in general rises gradually after a meal and returns to its premeal status in about two hours \cite{daly1998acute}. Therefore, as a usual fact, the blood glucose status $S^{i}_{n,t}|_{t=12}$ and $S^{i}_{n,t}|_{t=18}$ for patient $u_{i}$ at 12 p.m. and 6 p.m. (time for lunch or dinner) can be similar, even though these two moments are not closely adjacent. 

To fully explore the correlations among nonadjacent status and mine heterogeneous correlation beyond homogeneous sequences, a graph of status is constructed to represent the correlative similarity in different clips under the global view. Specifically, for each user/patient $u_i$ with historical status set $\{S^{i}_{\cdot t}\}_{t=1,\cdots,|T|}$, we define the temporal correlation graph as $G(V_i,E_i,g_i)$, where $V_i=\{S^{i}_{\cdot t}\}_{t=1,\cdots,|T|}$ denotes the nodes of the graph, $E_i=V_i\times V_i$ is the set of edges, and $g_i\in\mathbb{R}^{T\times T}$ represents the graph adjacency matrix with its element $g_i{[t_1,t_2]}\in[0,1]$ being the normalized cosine similarity between nodes $S^i_{\cdot t_1}$ and $S^i_{\cdot t_2}$. 

\begin{algorithm}[t]
	\caption{Hierarchical Global Views-guided
Sequence Representation Learning : $\tilde{y}_{i}$=HGV($U$, $S$, $F_{b}$, $y$, $T$ )}
	\label{a1}
	\begin{algorithmic}[1]
		\REQUIRE ~~\\
		$U$: User/patient set, $S$: Dynamic information tensor, $F_{b}$: Static information matrix, $y$: Historical risk probability vector, $T$: Time steps
		
		\ENSURE ~~ \\
		$\tilde{y}_{i}$: Risk probability for user/patient $u_{i}$
		
		\WHILE{$u_{i}\in U$}
		\FOR{$S^{i} \in S$} 
		\STATE
		$g_{i} \leftarrow TemporalCorrelationGraph(S^{i},T)$
		\STATE
		$E_{g}(g_{i}) \leftarrow GlobalGraphEmbeding(g_{i})$
		\ENDFOR
		\STATE
		$E_{b}(f_{b}^{i}) \leftarrow EmbeddingNet(f_{b}^{i})$
		\STATE
		$G^{i}  \leftarrow FuseNet(Concat(E_{g}(g_{i}),E_{b}(f_{b}^{i})))$
		\FOR{$n \in \{1,2,\ldots,N_{d}\}$} 
		\STATE
		$h_{n\cdot}^{i} \leftarrow LSTMs(S_{n\cdot}^{i},T)$ //Eq.(\ref{e3})
		\STATE
		$\alpha_{n\cdot}^{i} \leftarrow Softmax(\beta$-$Attn(S_{n\cdot}^{i},T))$ //Eq.(\ref{b-attn}-\ref{e8})
		\STATE
		$E_{d}(S_{n\cdot}^{i})\leftarrow [h_{n\cdot}^{i}] \cdot [\alpha_{n\cdot}^{i}]^\mathrm{T}$ //Eq.(\ref{e9})
		\ENDFOR
		\STATE
		$H^{i} \leftarrow MutilHead([E^{i}_{d},G^{i}])$ //Eq.(\ref{MultiHead})
		\STATE $\tilde{y_{i}} \leftarrow F(H^{i})$ //Eq.(\ref{AGG}-\ref{MLP})
		\ENDWHILE
	\end{algorithmic}
\end{algorithm}

\subsubsection{\textbf{Global Graph Embedding}}
Different from the conventional unordered graph, the temporal correlation graph $g_i$ is constructed in a temporal order as shown in Fig. \ref{fig2}, it means that the convolution neural network can be directly applied to $g_i$ using a clip aware sliding convolution kernel, thus to obtain a global embedding of temporal correlation graph $g_i$. One thing worth pointing out is that the traditional used GNNs \cite{DBLP:conf/iclr/KipfW17}, such as GCN and GAT, are only feasible to graph node representation rather than the graph representation itself, \emph{i.e.}, global graph representation. Specifically, let $W_{i}^{l}$ and $b_{i}^{l}$ be the convolution kernel parameters of layer $l$ in 2-D convolutional networks, the output $g_{i}^{(l)}$ at each layer $l\in \{1,\cdots,L\}$ is given as follows:
\begin{equation}
g_{i}^{(l)}=f(g_{i}^{(l-1)} \star W_{i}^{l} + b_{i}^{l}),
\end{equation}
in which $\star$ is the convolutional operation, and $f(\cdot)$ denotes a non-linear activation function (ReLU \cite{DBLP:journals/jmlr/GlorotBB11}) is used in our case. With the sliding of convolution kernel, the clip-aware patterns encoded in temporal status correlation can be extracted effectively. Following the convolution neural network is a fully connected layer to is to obtain the final global graph embedding $E_g(g_{i})$ and we have: 
\begin{equation}
E_g(g_{i})=f(W_{i}^{FC}\cdot g_{i}^{(L)} + b_{i}^{FC}),
\end{equation}
where $W_{i}^{FC}$ and $b_{i}^{FC}$ are parameters of the fully connected layer, and $g_{i}^{(L)}$ denotes the concatenated vector of the output of the last layer of convolution neural network.

\subsection{Harmonic $\beta$-attention}
\subsubsection{\textbf{Temporal Modeling at Channel Level}}
To capture the temporal dependencies within an individualized sequential signal at channel level, we set the Long Short-Term Memory networks (LSTMs) \cite{DBLP:journals/neco/HochreiterS97} as the backbone. Specifically, one of time series channels $S_{n\cdot}^{i}=(S_{n,1}^{i},S_{n,2}^{i},\ldots,S_{n,T}^{i}) \in S^{i}$ is fed into a LSTM network and the output for feature $S_{n\cdot}^{i}$ at time $t$ can be obtained by:
\begin{equation}
h_{n,1}^{i},h_{n,2}^{i},\ldots,h_{n,T}^{i}=LSTM_{\Theta}^{(n)}(S_{n,1}^{i},S_{n,2}^{i},\ldots,S_{n,T}^{i}),
\label{e3}
\end{equation}
where $h_{n,t}^{i} \in \mathbb{R}^{d_{1}}$ is the hidden representation for $S_{n,t}^{i}$ and $\Theta$ is the parameter space need to be learned for each LSTM network.

\subsubsection{\textbf{$\beta$-Attn: An Adaptive Key-query Attention}}
Learning the trends of several important markers effectively is deemed as a primary challenge in risk prediction task, especially in the healthcare and financial domains. Existing studies \cite{DBLP:conf/aaai/MaZWRWTMGG20,DBLP:conf/kdd/BaytasXZWJZ17} have proved the effectiveness of considering time-aware decay in long sequences and capturing the irregularity among different visit records. However, some significant variable values in specific visit record should be given more weights than those less significant values in nearer ones. For example, patients with severe diabetes in ICU may experience sudden blood glucose spikes and gradually recover as the result of therapeutic interventions. Obviously, such abnormal values deserve more attention \cite{park2013predicting}, which calls for a global trade-off between time-aware decay and observation significance. The time-aware decay $d_{\omega}^{i}$ and the observation significance $o_{n,t}^{i}$ are defined as:
\begin{equation}
d_{\omega}^{i} = 1-\frac{\Delta t}{\max(\Delta t)},  
\end{equation}
\begin{equation}
o_{n,t}^{i} = \frac{\sigma(S_{n,t}^{i})}{\sigma(\max(|S_{n\cdot}^{i}|))},
\end{equation}
where $\Delta t$ is the time interval from time $t$ to the latest observation time $T$, $\max(\Delta t)=T$, and $\sigma(\cdot)$ the non-linear mapping function $sigmoid$. Furthermore, inspired by the F-score for model performance evaluation, which is a harmonic mean of precision and recall, namely the reciprocal of the average of the reciprocal of precision and the reciprocal of recall \cite{DBLP:journals/corr/abs-2010-16061}, we learn a global trade-off between time-aware decay and observation significance adaptively. Specifically, the trade-off between $d_{\omega}^{i}$ and $o_{n,t}^{i}$ is measured as:
\begin{equation}
\begin{aligned}
\beta_{n,t}^{i}&=\frac{1}{\frac{1}{\beta +1}\cdot \frac{1}{d_{\omega}^{i}}+\frac{\beta}{\beta +1}\cdot \frac{1}{o_{n,t}^{i}}}
\\& =(1+\beta) \cdot \frac{d_{\omega}^{i}\cdot o_{n,t}^{i}}{\beta\cdot d_{\omega}^{i} + o_{n,t}^{i}},
\end{aligned}
\end{equation}
where $\beta$ is the trainable trade-off parameter. Furthermore, partly following the manner of \cite{DBLP:conf/aaai/MaZWRWTMGG20}, we define attention weights for $\beta$-Attn as:
\begin{equation}
\begin{aligned}
    &\theta_{n,t}^{i}= 
    \\&\tanh(\frac{(W_{n}^{q}\cdot h_{n,T}^{i}  )^T\cdot  W_{n}^{k}\cdot h_{n,t}^{i}}
   {\gamma_{n}^{i} \cdot log(c+(1-\sigma((W_{n}^{q}\cdot h_{n,T}^{i}  )^T\cdot  W_{n}^{k}\cdot h_{n,t}^{i}))) \cdot \beta_{n,t}^{i} \cdot T}),
\end{aligned}
\label{b-attn}
\end{equation}
where $\gamma_{n}^{i}$ is also a parameter that needs to be learned, $c$ represents a constant, and $W_{n}^{q} \in \mathbb{R}^{d_{2}\times d_{1}}$, $W_{n}^{k} \in \mathbb{R}^{d_{2}\times d_{1}}$ are projection matrices to map the query and key vectors in key-query attention, respectively. Finally, the normalized attention weights can be obtained by:
\begin{equation}
\alpha_{n,1}^{i},\alpha_{n,2}^{i},\ldots,\alpha_{n,T}^{i}=softmax(\theta_{n,1}^{i},\theta_{n,2}^{i},...,\theta_{n,T}^{i}).
\label{e8}
\end{equation}
Based on Eq. \ref{e8}, the weighted channel-wise representation $E_d(S_{n\cdot}^{i}) \in \mathbb{R}^{d_{1}}$ for the channel signal $S_{n\cdot}^{i}$ can be calculated by:
\begin{equation}
E_{d}(S_{n\cdot}^{i})=[h_{n\cdot}^{i}] \cdot [\alpha_{n\cdot}^{i}]^\mathrm{T}.
\label{e9}
\end{equation}

\subsection{Heterogeneous Information Aggregation}
Let $E_{b}(f_{b}^{i})$ be the embedding for the static information $f_{b}^{i}$ of the $i$-th instance, and then, both the $E_{b}(f_{b}^{i}) \in \mathbb{R}^{d_{b}}$ and the global graph embedding $E_{g}(g_{i}) \in \mathbb{R}^{d_{g}}$ are concatenated to obtain a fused instance level representation $G_{i}\in \mathbb{R}^{d_{1}}$ via a linear Fusenet (a one-layer MLP). Furthermore, by stacking the instance level representation $G_{i}$ and the channel-level representations $E^{i}_{d}=[E_{d}(S_{n\cdot}^{i})]_{n=1,\cdots, N_d} \in \mathbb{R}^{d_{1}\times N_d}$, we have hierarchical representations $E_{i}=[E^{i}_{d},G^{i}] \in \mathbb{R}^{d_{1}\times (N_d+1)}$ for the $i$-th instance. 

To capture the interdependencies among these hierarchical representations learned from both instance level and channel level, a strategy of multi-head attention \cite{DBLP:conf/nips/VaswaniSPUJGKP17} is adopted. To be specific, given the hierarchical stacked representations $E_{i}$ as the input, let $head_{i}^{(h)}$ be the embedding through the $h$-th attention head given by:
\begin{equation}
    \begin{aligned}
        head_{i}^{(h)}(E_{i})= softmax(\frac{Q^{(h)}_{i}\cdot K_{i}^{(h)T}}{\sqrt{d_{1}}}) \cdot V^{(h)}_{i},
    \end{aligned}
\label{attn}
\end{equation}
\begin{equation}
\begin{aligned}
    Q_{i}^{(h)}&= W_{i}^{Q^{(h)}}\cdot E_{i}, \\
    K_{i}^{(h)}&= W_{i}^{K^{(h)}}\cdot E_{i}, \\
    V_{i}^{(h)}&= W_{i}^{V^{(h)}}\cdot E_{i}.
\end{aligned}
\end{equation}
where $\{Q_{i}^{(h)}, K_{i}^{(h)}, V_{i}^{(h)}\in \mathbb{R}^{ d_{1}\times (N_d+1)}\}$ are the query, key, and value matrices, respectively, for the $h$-th head, and $\{W_{i}^{Q^{(h)}},W_{i}^{K^{(h)}}, W_{i}^{V^{(h)}} \in \mathbb{R}^{ d_{1}\times d_{1}}\}$ are the corresponding projection matrices. To further integrate the multi-head embeddings $head_{i}^{(h)}(E_{i})$, $h=1,\cdots, N_H$, we have:
\begin{equation}
\begin{aligned}
H^{i} &= MultiHead(E_{i}) \\&= (W^{H})^T\cdot Concat(head_{i}^{(1)}(E_{i}), \ldots, head_{i}^{(N_H)}(E_{i})),
\label{MultiHead}
\end{aligned}
\end{equation}
where $H^i\in \mathbb{R}^{d_1\times (N_d+1)}$ and $W^{H}\in \mathbb{R}^{(d_1\times N_H) \times d_1}$ is a linear projection matrix.

\begin{figure}[t]
	\centering
	\includegraphics[width=0.4\textwidth,height=3.5cm]{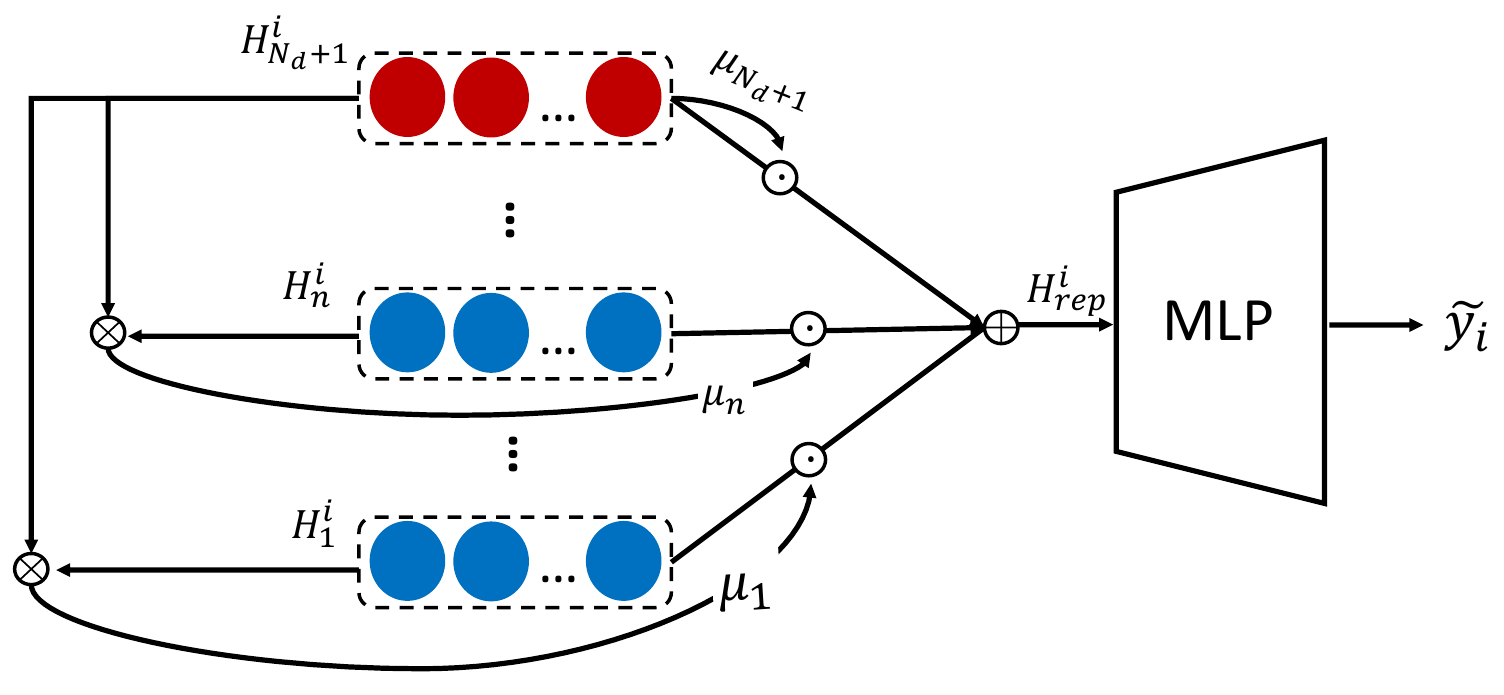}
	\caption{Pipeline of heterogeneous information aggregation.
	}
	\label{aggregation}
\end{figure}

Clearly, as a hierarchical description of the instance $i$, the first $N_d$ columns of $H^i$ reflect indeed the characterization of instance $i$ at the channel level, whereas the last one $H^i_{N_d+1}$ reflects the global view on it at the instance level. Intuitively, in a real risk prediction task, the channel level characterizations of dynamic statuses should be consistent as far as possible with the global view with embedded static information and temporal correlation graph. It means we can obtain a unified representation with multi-granularity information via the guidance of global view. Specifically, we have the unified representation $H^i_{rep} \in \mathbb{R}^{d_1}$ as follows:
\begin{equation}
    H^i_{rep}=\sum\limits_{n=1}^{N_d} \underbrace{\mu_{n} \cdot H_{n}^{i}}_{channel-level} + \underbrace{\mu_{N_d+1}\cdot H_{N_d+1}^{i}}_{instance-level}
    \label{AGG}
\end{equation}
where $\{\mu_{m}=softmax((H_m^i)^T \cdot H_{N_d+1}^i)\}_{m=1,\cdots,N_{d}+1}$ denote the global view guided weights on representations at both channel level and instance level. In fact, it is not hard to see, the more relevant between the channel level characterization to the global guidance at the instance level, the larger the weight will be confirmed. Finally, the risk probability $\hat{y_{i}}$ can be obtained via an MLP predictor:
\begin{equation}
\hat{y_{i}}=MLP(H^i_{rep}),
\label{MLP} 
\end{equation}
The pipeline of the heterogeneous information aggregation is given in Fig. \ref{aggregation}.

\subsection{Model Optimization}
\subsubsection{\textbf{Model Regularization}}
With the practical experience \cite{DBLP:conf/aaai/MaZWRWTMGG20,DBLP:conf/ijcai/ChuLWWWG19}, the DeCov loss is widely introduced for obtaining non-redundant representations by minimizing the cross-covariance of hidden activation \cite{DBLP:journals/corr/CogswellAGZB15}. Specifically, to capture the coupling among multi-head attention layers, the covariances between all pairs of activation $i$ and $j$ from matrix $C$ can be calculated as:
\begin{equation}
C_{i,j}=\frac{1}{B}\sum\limits_{b} (H_{b}^{i}-\frac{1}{B}\sum\nolimits_{b} H_{b}^{i})(H_{b}^{j}-\frac{1}{B}\sum\nolimits_{b} H_{b}^{j}),
\end{equation}
where $B$ is the batch size, $H_{b}^{i}$ the $n$-th refined representation for the $b$-th case in the batch (for convenience, $n$ is not distinguished here). To capture similar dependencies among different heads, we need to minimize covariance by penalizing the norm of $C$. However, the diagonal of $C$ should not be minimized as well as the norm due to the fact that it measures dynamic range of activation, which has nothing to do with our purpose. Thus, the regularization loss can be defined as:
\begin{equation}
\mathcal{L}_{DeCov}=\frac{1}{2}(\Vert C \Vert_{F}^{2}-\Vert diag(C) \Vert_{2}^{2}),
\end{equation}
where $\Vert \cdot \Vert$ is the Frobenius norm and the $diag(\cdot)$ is the operator of extracting a vector from the main diagonal of a matrix. Moreover, to avoid over-fitting, we have also deployed some strategies such as residual connection and dropout operation \cite{DBLP:conf/cvpr/HeZRS16}.

\subsubsection{\textbf{Risk Prediction}}
Generally, the risk prediction task is defined as a binary classification problem, where an observed risky endpoint status is assigned a target value 1, otherwise 0. Specifically, we use the cross entropy as the loss functions for classification,
\begin{equation}
	\mathcal{L}_{C}=-\left(\sum\limits_{y_{i}\in R^{+}}\log\widehat{y}_{i}+\sum\limits_{y_{i}\in R^{-}} \log(1-\widehat{y}_{i})\right),
\end{equation}
in which $R^{+}$ and $R^{-}$ are the records with positive and negative target values. 
Finally, to both capture the coupling among multi-head attention layers and improve the performance of risk prediction, the task can be solved by minimizing the following hybrid loss 
\begin{equation}
\mathcal{L}=\mathcal{L}_{C}+\lambda_{d}\cdot \mathcal{L}_{DeCov}.
\end{equation}
where $\lambda_{d}$ is a trade-off parameter.

In summary, the implementation details about the proposed HGV are outlined in Algorithm \ref{a1}.

\section{Experimental Results and Analysis }
In this section, we aim to answer the following research questions:
\begin{itemize}
	\item {RQ1: How to set the risk prediction on both healthcare and financial tasks?}
	\item {RQ2: How to evaluate the performance of the HGV?}
	\item{RQ3: How does the HGV perform and why?}
\end{itemize}

\subsection{Data Description and Settings (RQ1)}
\subsubsection{\textbf{Risk Prediction on Healthcare Data}}
Our healthcare risk prediction task is based on \textbf{MIMIC-III} \cite{johnson2016mimic}, a large publicly available benchmarking dataset. In this task, the mortality risk needs to predict is a primary outcome of interest in acute care, which is the key to improving outcomes for those at-risk patients. Specifically, following the preprocessing pipeline established by the \cite{Harutyunyan2019}, we split the MIMIC-III dataset into train, val and test set, and the detailed statistics has been summarized in Table \ref{t2}. 

\begin{itemize}
	\item{\textbf{MIMIC-III Dataset\footnote{https://mimic.physionet.org/}:} 
	\underline{M}edical \underline{I}nformation \underline{M}art for \underline{I}ntensive \underline{C}are is a large, single-center database comprising information relating to patients admitted to critical care units at a large tertiary care hospital. Data includes vital signs, medications, laboratory measurements, observations, and notes charted by care providers, fluid balance, procedure codes, diagnostic codes, imaging reports, hospital length of stay, survival data, and more. Noted that one patient may have more than one record. 
	}
\end{itemize}

\begin{table}[t]   
	\renewcommand\arraystretch{1.5}
	\centering
	\caption{Detailed statistics of MIMIC-III and Ant Group-MYBank Datasets. (sparsity is the proportion of positive samples, indicating the imbalanced and skewed level of dataset).} 
	\begin{tabular}{p{3cm}<{\centering}|p{1.4cm}<{\centering}|p{2.6cm}<{\centering}} 
		\hline  
		Data Sources & MIMIC-III & Ant Group-MYBank \\ 
		\hline
		\#patients/users (SMEs) &18,094 &7,947\\  
		\#time steps $T$ &48 &14 \\
		\#Num. of static info. $N_b$ & 7 &9 \\
		\#Num. of dyn. info. $N_d$  & 17 &26 \\
		\#sparsity &0.6138 &0.9189\\ 
		\hline
		\#train samples &14,681 &5,564\\
		\#valid samples &3,222 &795\\
		\#test samples &3,236 &1,588\\
		\hline
	\end{tabular} 
	\label{t2} 
\end{table}

\subsubsection{\textbf{Risk Prediction on Financial Data.}}
The detailed statistics for the real-world industrial dataset are also given in Table \ref{t2}. Noted, as a SMEs credit overdue risk prediction task, it is more \textbf{imbalanced} and \textbf{skewed} than MIMIC-III, which brings more challenges.
\begin{itemize}
	\item{\textbf{Ant Group-MYBank Dataset\footnote{https://www.mybank.cn/}:}} 
    It includes rich personal profiles and loaning behavior data of SMEs owners of MYBank such as the age, gender, education and loan amount, the current balance, the duration time to the latest loan, the number of loans and so on. 
\end{itemize}
We randomly collect about 484,828 traffic logs across one month (e.g., from Dec. 1, 2021 to Jan.1, 2022) from an online SMEs loan scenario in MYBank, Ant Group. We gather the repayment feedback of each SME owner as the target risk status. Note that all above data are definitely authorized by the SME users since they hope to apply for loan in our bank, and they should provide their lending history and personal profiles.

\begin{table*}[t]
    \renewcommand\arraystretch{1.5}
	\centering
	\caption{Performance comparison on public benchmark healthcare risk prediction task (the best is in \textcolor{red}{red}, the second is in \textcolor{blue}{blue}, the third is in \textcolor{green}{green} and the improvements are in bracket).}
	\begin{tabular}{|p{4.2cm}<{\centering}|p{4cm}<{\centering}|p{4cm}<{\centering}|p{4cm}<{\centering}|}
		\hline
		\hline
		\multirow{3}{*}{Model}
		&\multicolumn{3}{c|}{Overall Performance of Healthcare Risk Prediction in Benchmark Task.}\\
		\cline{2-4}
		&\multicolumn{3}{c|}{MIMIC-III Dataset (Bootstrapping = 1000)}\\
		\cline{2-4}
		&AUROC &AUPRC &min(Se, P+)\\
		\cline{2-4}
		\hline
		LR \cite{DBLP:journals/ml/YuHL11}	&$0.8485\pm0.010$ (-2.33\%)	&$0.4758 \pm 0.028$ (-6.28\%)	&$0.4643\pm 0.022$ (-5.66\%)\\
		GBDT \cite{friedman2001greedy} &$0.8468 \pm 0.011$ (-2.50\%)	&$0.5032 \pm 0.027$ (-3.54\%)	&$0.4916 \pm 0.022$ (-2.93\%)\\
		\hline
		Attn-GRU \cite{DBLP:conf/emnlp/ChoMGBBSB14} &$0.8628\pm 0.011$ (-0.90\%)	&$0.4989 \pm 0.022$ (-3.97\%)	&$0.5026 \pm 0.028$ (-1.83\%)\\
		RETAIN \cite{DBLP:conf/nips/ChoiBSKSS16} &$0.8313\pm0.014$ (-4.05\%)	&$0.4790\pm 0.020$ (-5.96\%)	&$0.4721\pm 0.022$ (-4.88\%)\\
		T-LSTM \cite{DBLP:conf/kdd/BaytasXZWJZ17} &$0.8628\pm 0.011$ (-0.90\%)	&$0.4989 \pm 0.022$ (-3.97\%)	&$0.5026 \pm 0.028$ (-1.83\%)\\
		MCA-RNN \cite{DBLP:conf/icdm/LeePJM18} &$0.8587 \pm 0.013$ (-1.31\%)	&$0.5003\pm0.028$ (-3.83\%)	&$0.4932\pm 0.024$ (-2.77\%)\\
		\hline
		Transformer \cite{DBLP:conf/nips/VaswaniSPUJGKP17} &$0.8535\pm0.014$ (-1.83\%)	&$0.4917\pm0.022$ (-4.69\%)	&$0.5000\pm 0.019$ (-2.09\%)\\
		SAnD \cite{DBLP:conf/aaai/SongRTS18} &$0.8382\pm0.007$ (-3.36\%)	&$0.4545\pm0.018$ (-8.41\%)	&$0.4885\pm 0.017$ (-3.24\%)\\
		ConCare \cite{DBLP:conf/aaai/MaZWRWTMGG20} &\textbf{\textcolor{blue}{0.8659 $\pm$ 0.009 (-0.59\%)}}	&\textbf{\textcolor{green}{0.5238 $\pm$ 0.027 (-1.48\%)}}	&\textbf{\textcolor{blue}{0.5077 $\pm$ 0.022 (-1.32\%)}}\\
		GRASP \cite{DBLP:conf/aaai/ZhangGMWWT21} &\textbf{\textcolor{green}{0.8635 $\pm$ 0.009 (-0.83\%)}} 	&\textbf{\textcolor{blue}{0.5246 $\pm$ 0.028 (-1.40\%)}} 	&\textbf{\textcolor{green}{0.5068 $\pm$ 0.028 (-1.41\%)}}\\ 
		\hline
		HGV & \textbf{\textcolor{red}{0.8718 $\pm$ 0.010}}	&\textbf{\textcolor{red}{0.5386 $\pm$ 0.028}}	&\textbf{\textcolor{red}{0.5209 $\pm$ 0.023}} \\
		\hline
		\hline
	\end{tabular}
	\label{t3}
\end{table*}

\subsection{Experimental Settings (RQ2)}
\subsubsection{\textbf{Evaluation Metrics}}
We use Area Under the Receiver Operating Characteristic (AUROC) curve, Area Under the Precision-Recall Curve (AUPRC) and the Minimum of Precision and Sensitivity Min(Se, P+) to evaluate the performance of the proposed model on both healthcare risk prediction task and financial credit overdue risk prediction task. In fact, it is a widely used evaluation metric group in this kind of binary risk prediction problem with such imbalanced and skewed data. 

\subsubsection{\textbf{Baselines}}
We compare our proposed HGV with both traditional \cite{DBLP:journals/ml/YuHL11,friedman2001greedy}, time-aware \cite{DBLP:conf/nips/ChoiBSKSS16,DBLP:conf/kdd/BaytasXZWJZ17}, attention-based \cite{DBLP:conf/emnlp/ChoMGBBSB14,DBLP:conf/icdm/LeePJM18,DBLP:conf/nips/VaswaniSPUJGKP17,DBLP:conf/aaai/SongRTS18,DBLP:conf/aaai/MaZWRWTMGG20} and knowledge-enhanced \cite{DBLP:conf/aaai/ZhangGMWWT21} baselines. Noted, in order to ensure the fairness of the comparison, we quote the reported results or reproduced results for each baseline from their original literature or open-source implementations and all the reproduced ones have been fine-tuned by grid-searching strategy. 
\begin{itemize}
	\item{\textbf{LR} \cite{DBLP:journals/ml/YuHL11}: 
	It is a classic logistic regression model. We use a more elaborate version of the hand-engineered features given in benchmark pipelines \cite{Harutyunyan2019}.
	}
	\item{\textbf{GBDT} \cite{friedman2001greedy}: 
	It is also a classic tree-based ensemble learning method. The hand-engineered features used are the same as the LR baseline.
	}
	\item{\textbf{Attn-GRU} \cite{DBLP:conf/emnlp/ChoMGBBSB14}: 
	It is an attention-based model, where we add the plain attention to a multichannel GRU. 
	}
	\item{\textbf{RETAIN} \cite{DBLP:conf/nips/ChoiBSKSS16}:
	It is a well-known attention-based model, which can explore both temporal relationships and variable significance by using two-level neural attention model.  
	}
	\item{\textbf{T-LSTM} \cite{DBLP:conf/kdd/BaytasXZWJZ17}:
	It is a time-aware model, it can handle data with time irregularities. In this paper, it is modified into a supervised learning model.
	}
	\item{\textbf{MCA-RNN} \cite{DBLP:conf/icdm/LeePJM18}:
	It is an attention-based model, which utilize an attention-based RNN and a conditional deep generative model for capturing the heterogeneity in time series data.
	}
	\item{\textbf{Transformer} \cite{DBLP:conf/nips/VaswaniSPUJGKP17}:
	It is a well-known baseline self-attention based model. A flatten layer and FFNs in the final step are used to make the risk prediction.
	}
	\item{\textbf{SAnD} \cite{DBLP:conf/aaai/SongRTS18}: 
	It is a self-attention based model, which applies a masked, self-attention mechanism, and uses positional encoding and dense interpolation strategies for incorporating temporal order. 
	}
	\item{\textbf{ConCare} \cite{DBLP:conf/aaai/MaZWRWTMGG20}: 
	It is one of the state-of-the-art models in risk prediction task, which combine a new time-aware attention and multi-head self-attention networks with a cross-head decorrelation loss.
	}
	\item{\textbf{GRASP}} \cite{DBLP:conf/aaai/ZhangGMWWT21}:
	It is a knowledge-enhanced predictor, which extracts a k-nearest neighbor graph clustering from similar samples, and then enhances the performance by introducing cluster centers embedding learned by GCNs \cite{DBLP:conf/iclr/KipfW17}. Noted, we use the reported SOTA backbone, GRASP+ConCare, as the baseline.
\end{itemize} 

\subsubsection{\textbf{Parameter Settings}}
There are some training parameters involved in HGV, \emph{i.e.}, learning rate $lr$ and batch size $B$. In particular, for the batch size $B$ and learning rate $lr$, we set $B=256,lr=0.001$ for MIMIC-III and $B=128,lr=0.001$ for Ant Group-MYBank dataset, respectively. In addition, there are also some other hyperparameters in the backbone modules, LSTMs and CNNs, \emph{i.e.}, the hidden size $d_{1}$, layer number $L$, channel number $\lambda$ and kennel size $C_{k}$ of the layer $l$, stride $C_{s}$ for the convolution operation in the CNN. For these hyperparameters, we set $L_{LSTM}=1$, $L_{CNN}=2$, $\lambda_{l=1}=64$, $\lambda_{l=2}=128$, $C_{k}$=3 and $C_{s}$=1. Moreover, for the  hyperparameters in multi-head attention networks and $\beta$-Attn, \emph{i.e.}, the number of head $N_H$, hidden size for two attention layers $d_{1},d_{2}$. Taking both  the efficiency and performance into account, the settings for these hyperparameters are: $d_{1}=64$, $d_{2}=32$ and $N_H=4$. Note that, in order to guarantee the optimal parameters in experiments, we conduct grid searches and set the optimal hyperparameters for both our model and other competitors. More implementation details for the HGV can be referred with the open source code, which has been available at the GitHub repository\footnote{https://github.com/LiYouru0228/HGV}.

\begin{figure*}[!h]
	\centering
	\begin{subfigure}{0.38\textwidth}
		\centering 
		\includegraphics[width=\textwidth,height=6.5cm]{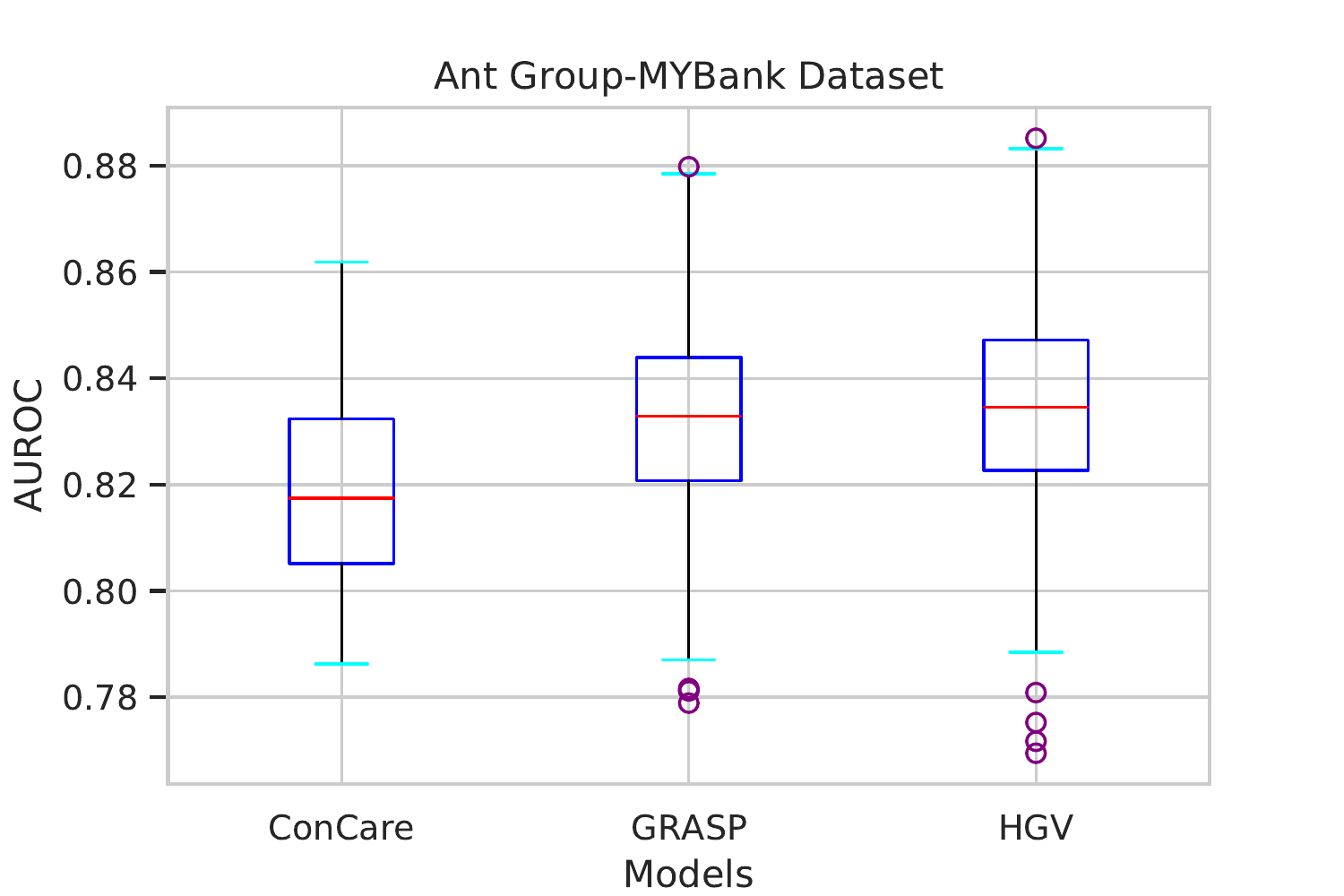} 
	\end{subfigure} 
	\begin{subfigure}{0.38\textwidth} 
		\centering \includegraphics[width=\textwidth,height=6.5cm]{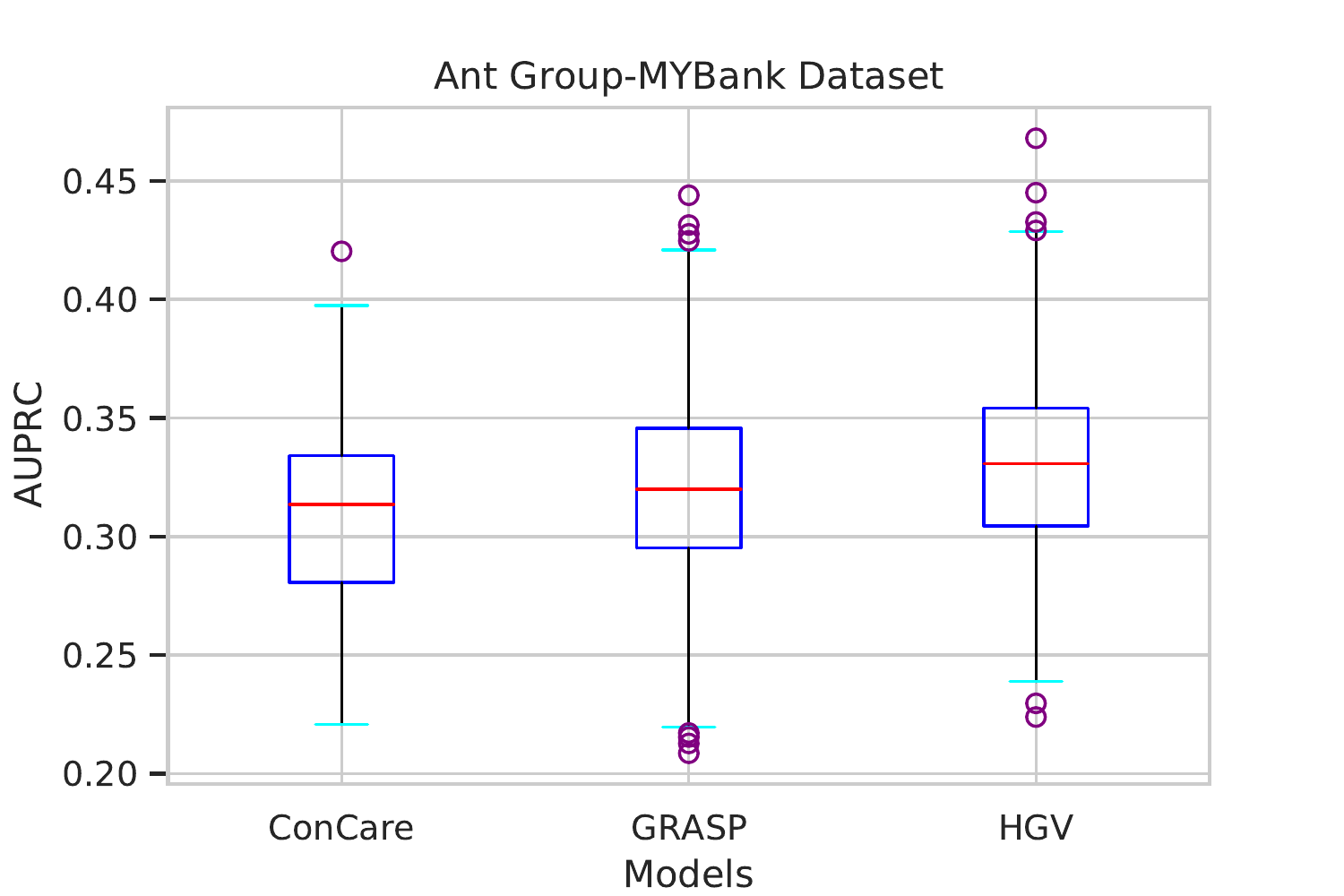} 
	\end{subfigure}
	\begin{subfigure}{0.38\textwidth}
		\centering 
		\includegraphics[width=\textwidth,height=6.5cm]{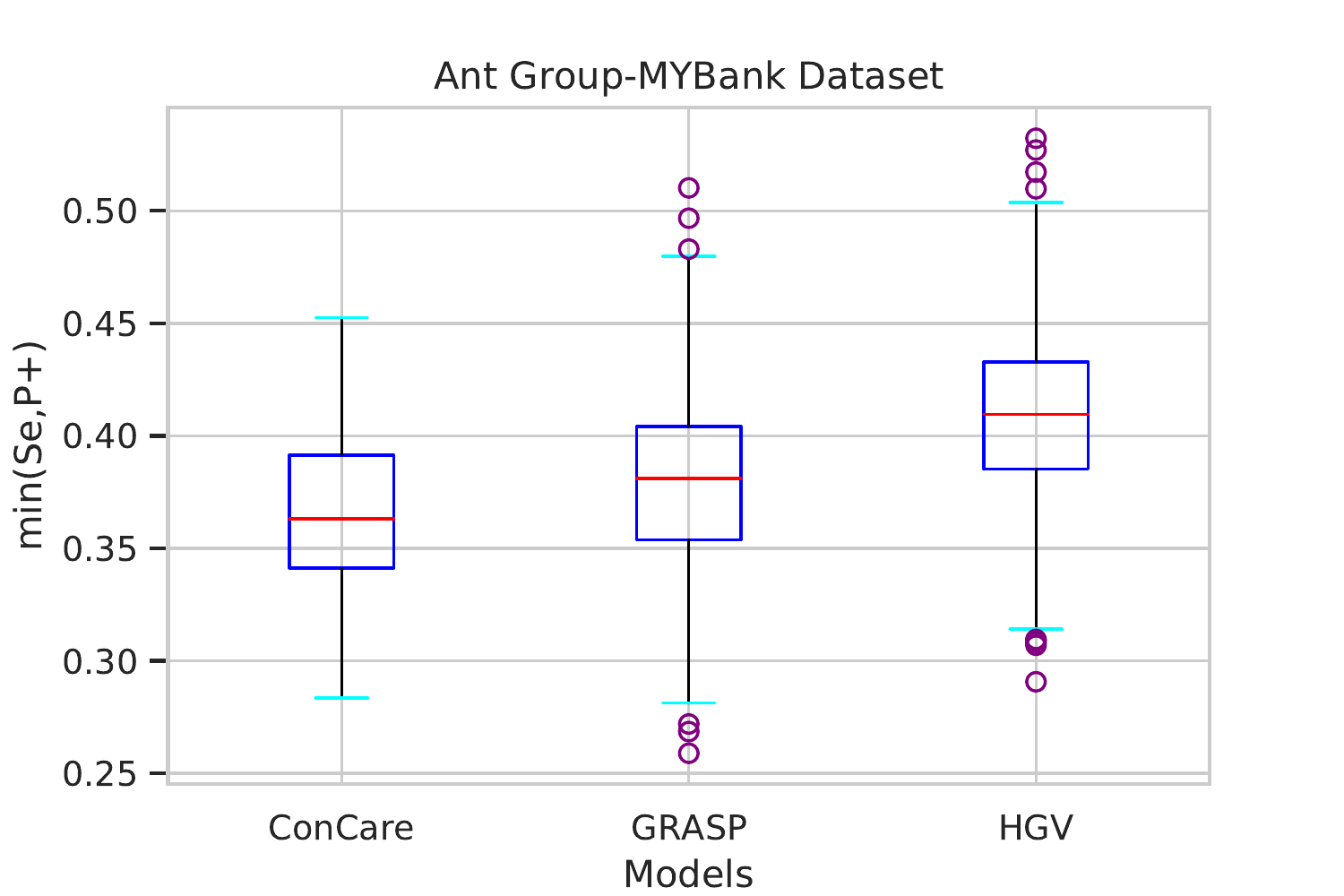} 
	\end{subfigure} 
	\caption{Performance comparison on the real-world industrial financial risk prediction task, between the proposed HGV with the strongest baselines in public benchmark task, ConCare and GRASP, respectively. (The box line diagrams have shown the middle value, 25\% and 75\% quantiles, minimum, maximum and outliers of evaluation metrics for each competitor. The confidence intervals and standard deviations are also estimated with bootstrapping on the data of each bucket for 1,000 times, which is the same as the setting on the Table \ref{t3}.)}
	\label{fig3}
\end{figure*}

\begin{figure*}[t]
	\centering
	\begin{subfigure}{0.32\textwidth}
		\centering 
		\includegraphics[width=\textwidth,height=5cm]{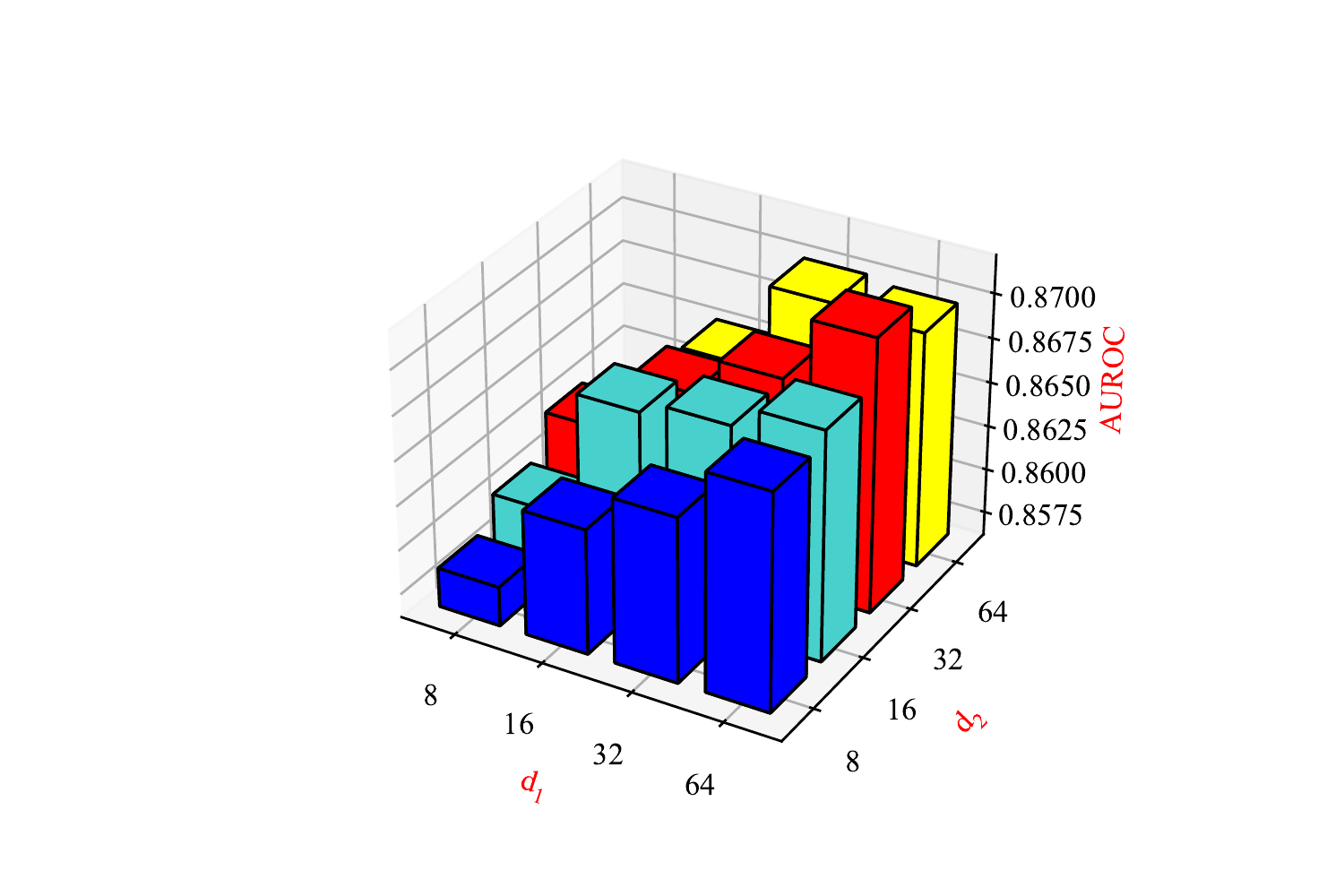} 
	\end{subfigure} 
	\begin{subfigure}{0.32\textwidth} 
		\centering \includegraphics[width=\textwidth,height=5cm]{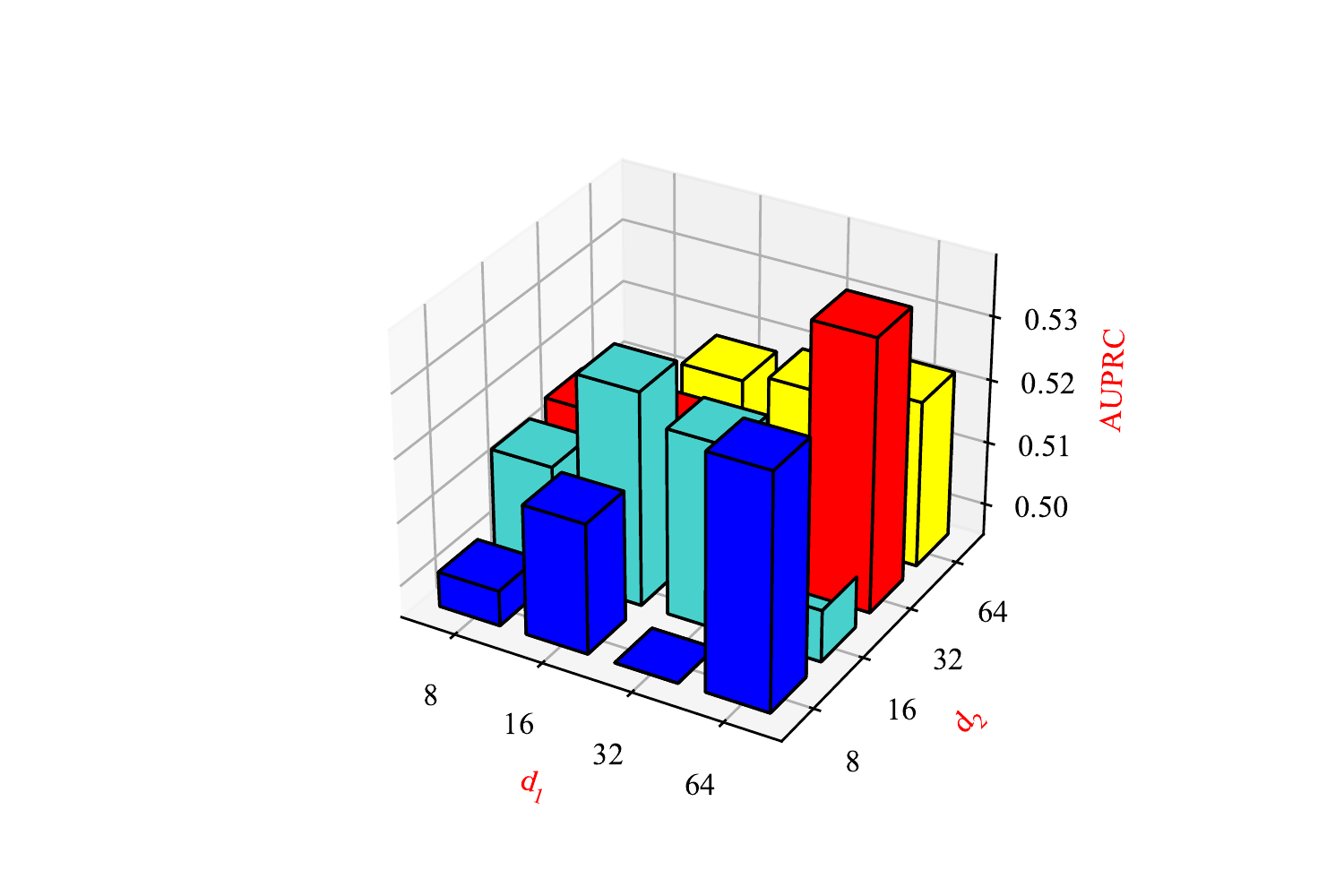} 
	\end{subfigure}
	\begin{subfigure}{0.32\textwidth}
		\centering 
		\includegraphics[width=\textwidth,height=5cm]{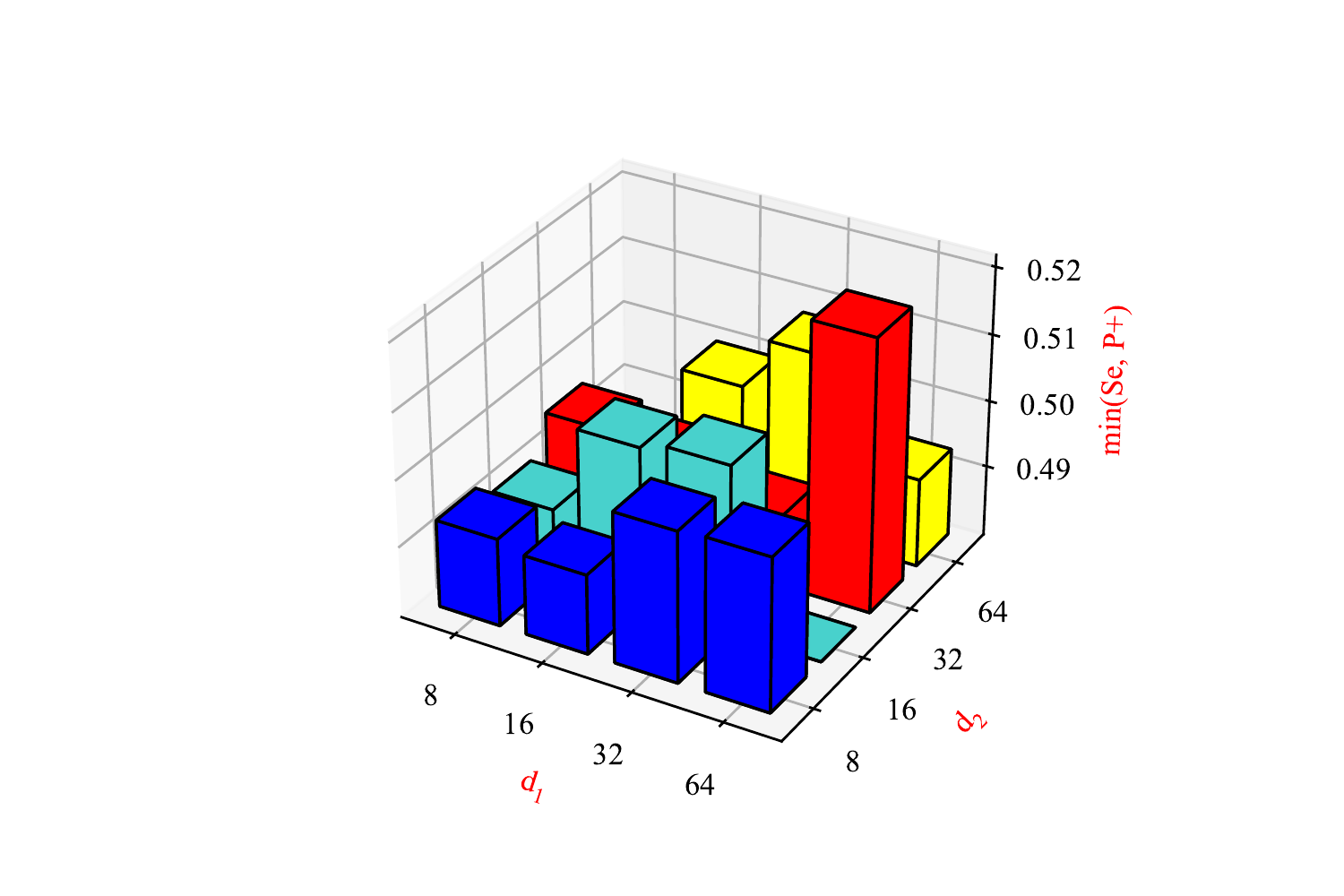} 
	\end{subfigure} 
	\caption{The performance under different hyper-parameter combinations: $d_{1} \times d_{2}$.}
	\label{pse1}
\end{figure*}
\begin{figure*}[t]
	\centering
	\begin{subfigure}{0.32\textwidth}
		\centering 
		\includegraphics[width=\textwidth,height=5cm]{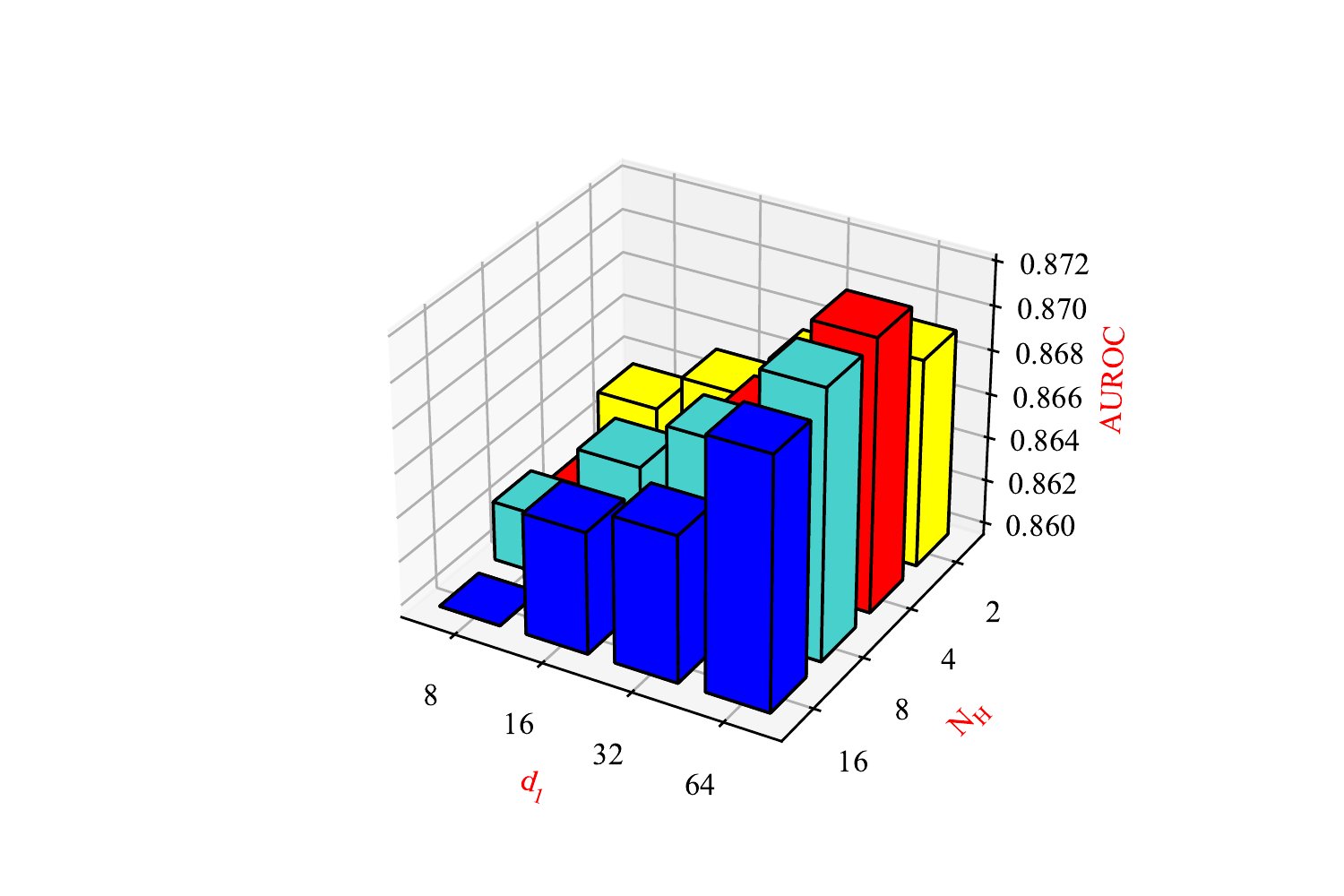} 
	\end{subfigure} 
	\begin{subfigure}{0.32\textwidth} 
		\centering \includegraphics[width=\textwidth,height=5cm]{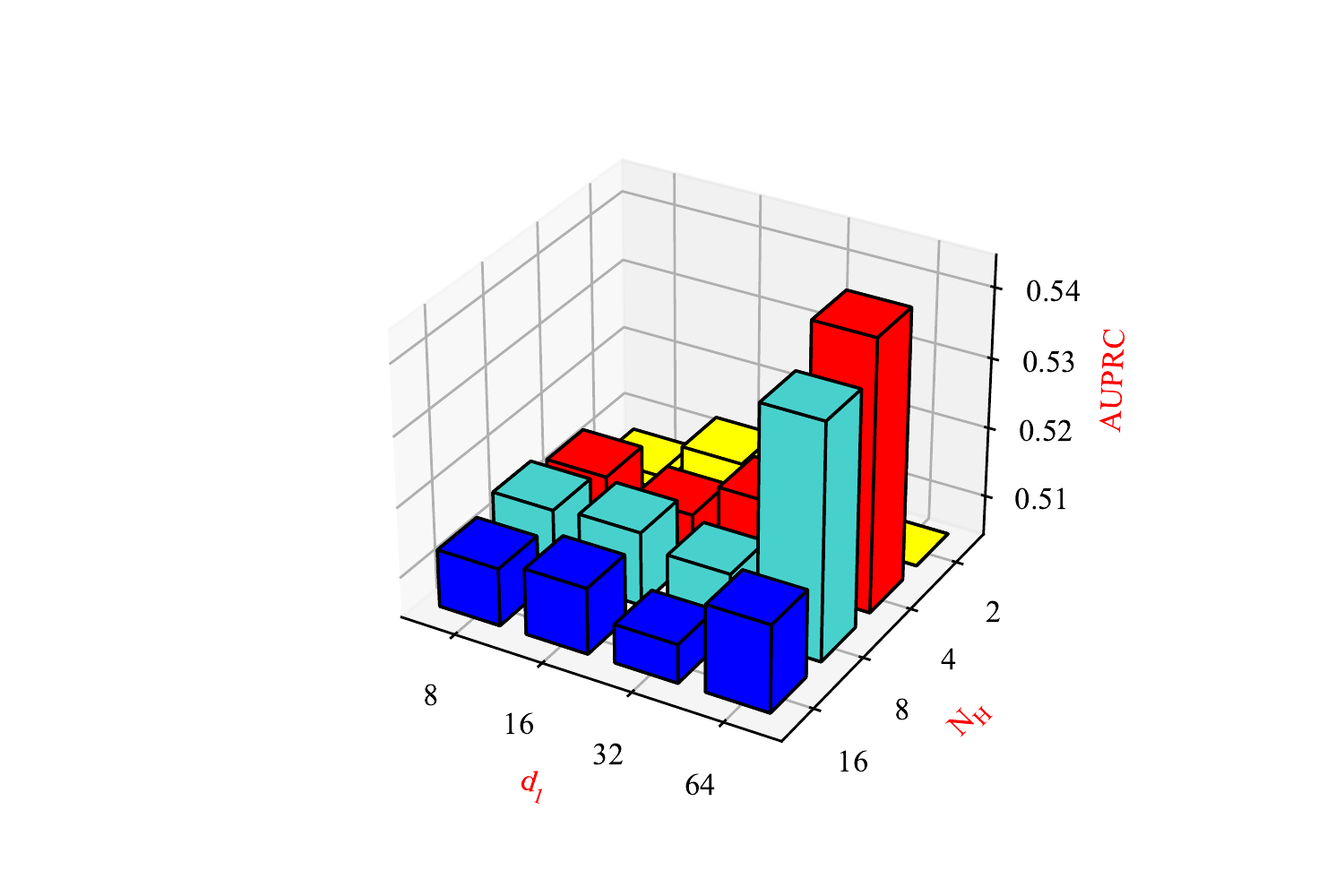} 
	\end{subfigure}
	\begin{subfigure}{0.32\textwidth}
		\centering 
		\includegraphics[width=\textwidth,height=5cm]{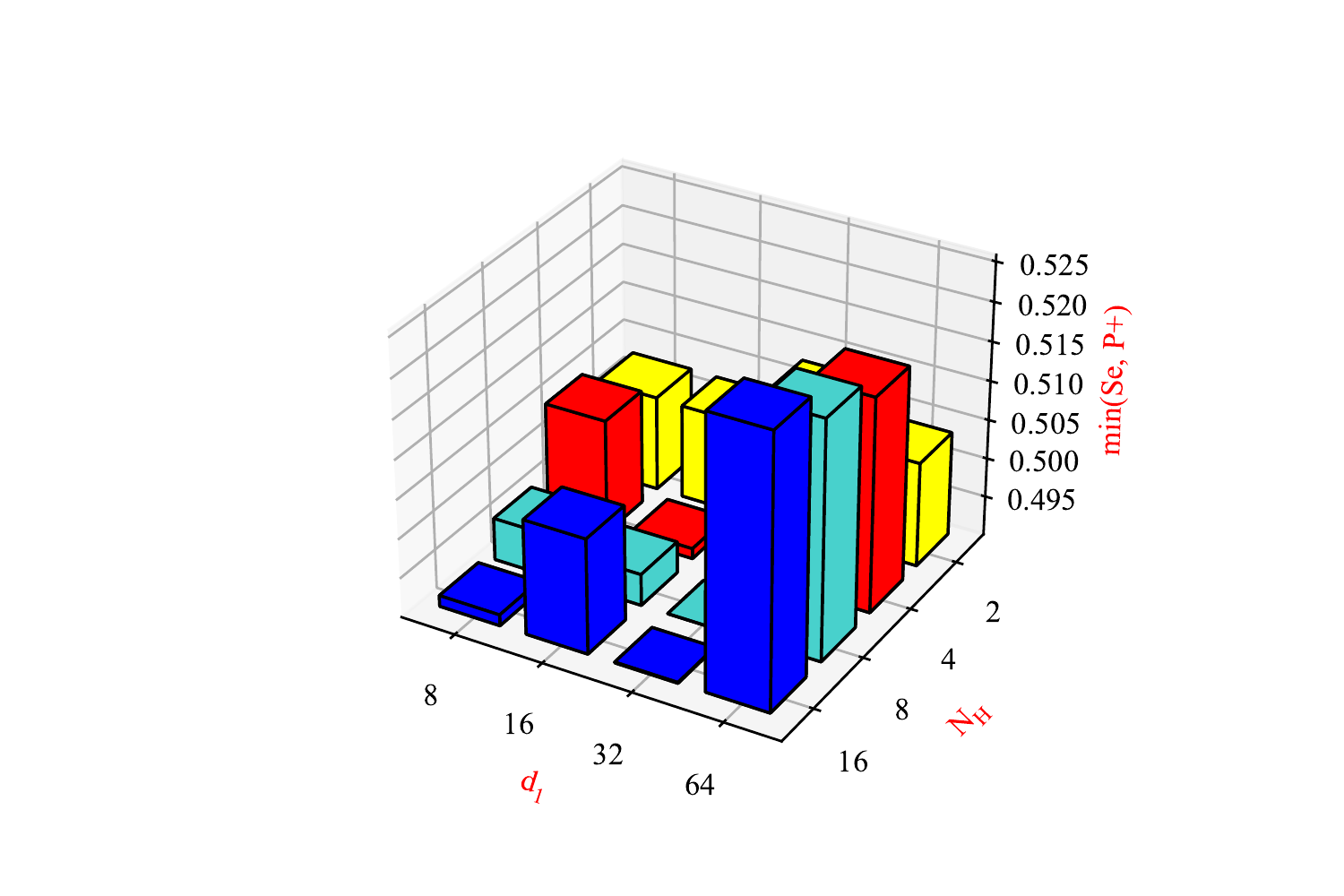} 
	\end{subfigure} 
	\caption{The performance under different hyper-parameter combinations: $d_{1} \times N_{H}$.}
	\label{pse2}
\end{figure*}
\begin{figure*}[t]
	\centering
	\begin{subfigure}{0.32\textwidth}
		\centering 
		\includegraphics[width=\textwidth,height=5cm]{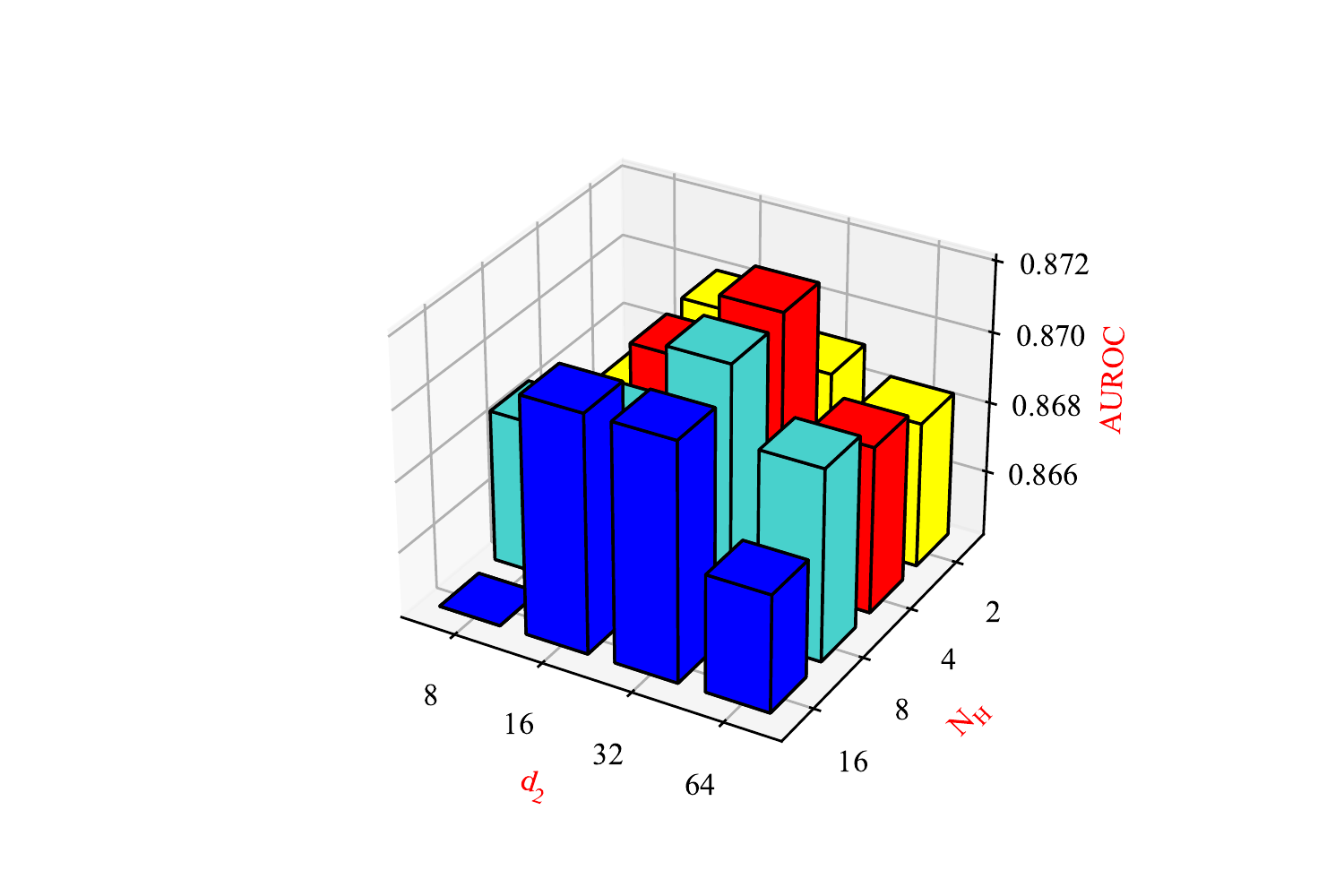} 
	\end{subfigure} 
	\begin{subfigure}{0.32\textwidth} 
		\centering \includegraphics[width=\textwidth,height=5cm]{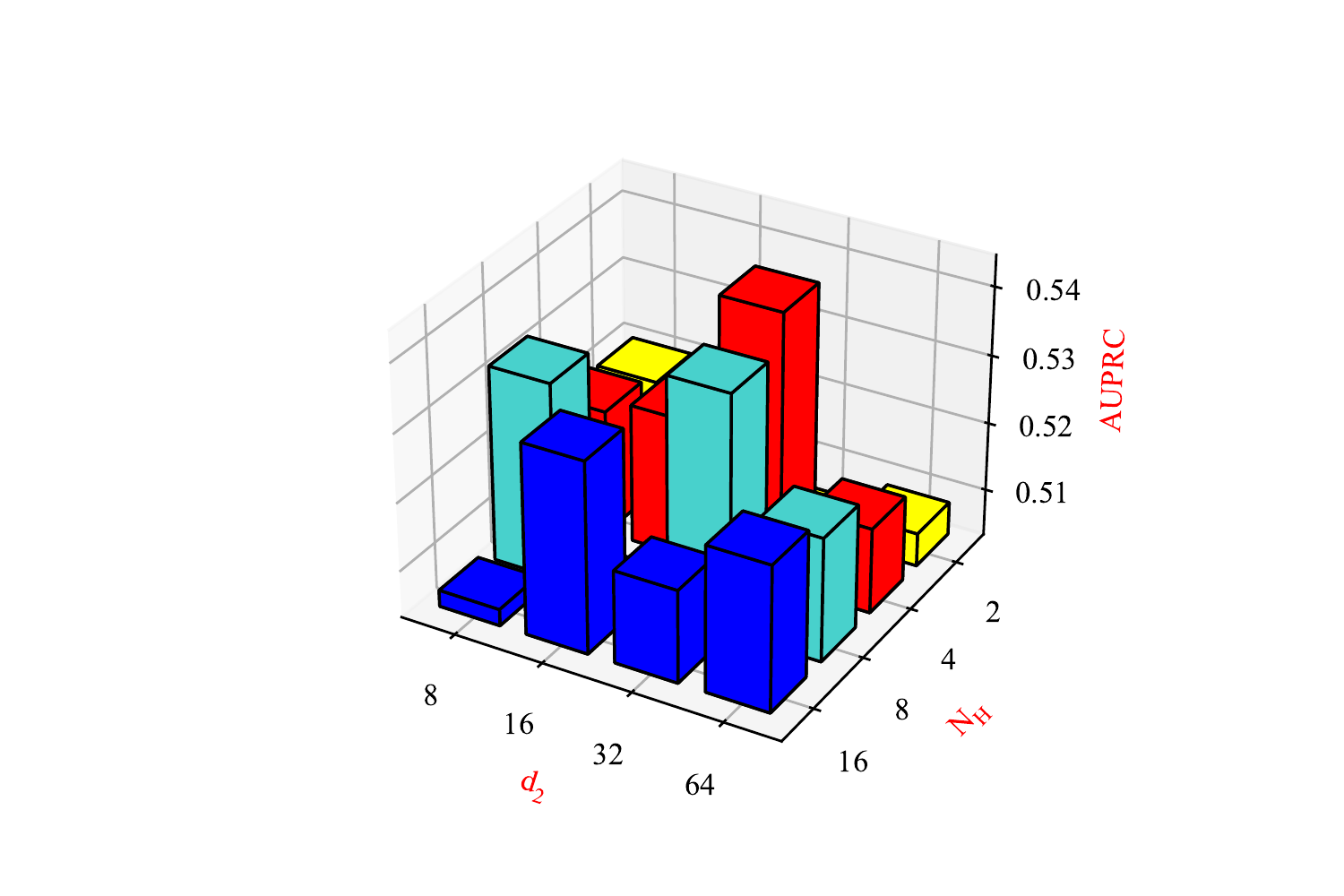} 
	\end{subfigure}
	\begin{subfigure}{0.32\textwidth}
		\centering 
		\includegraphics[width=\textwidth,height=5cm]{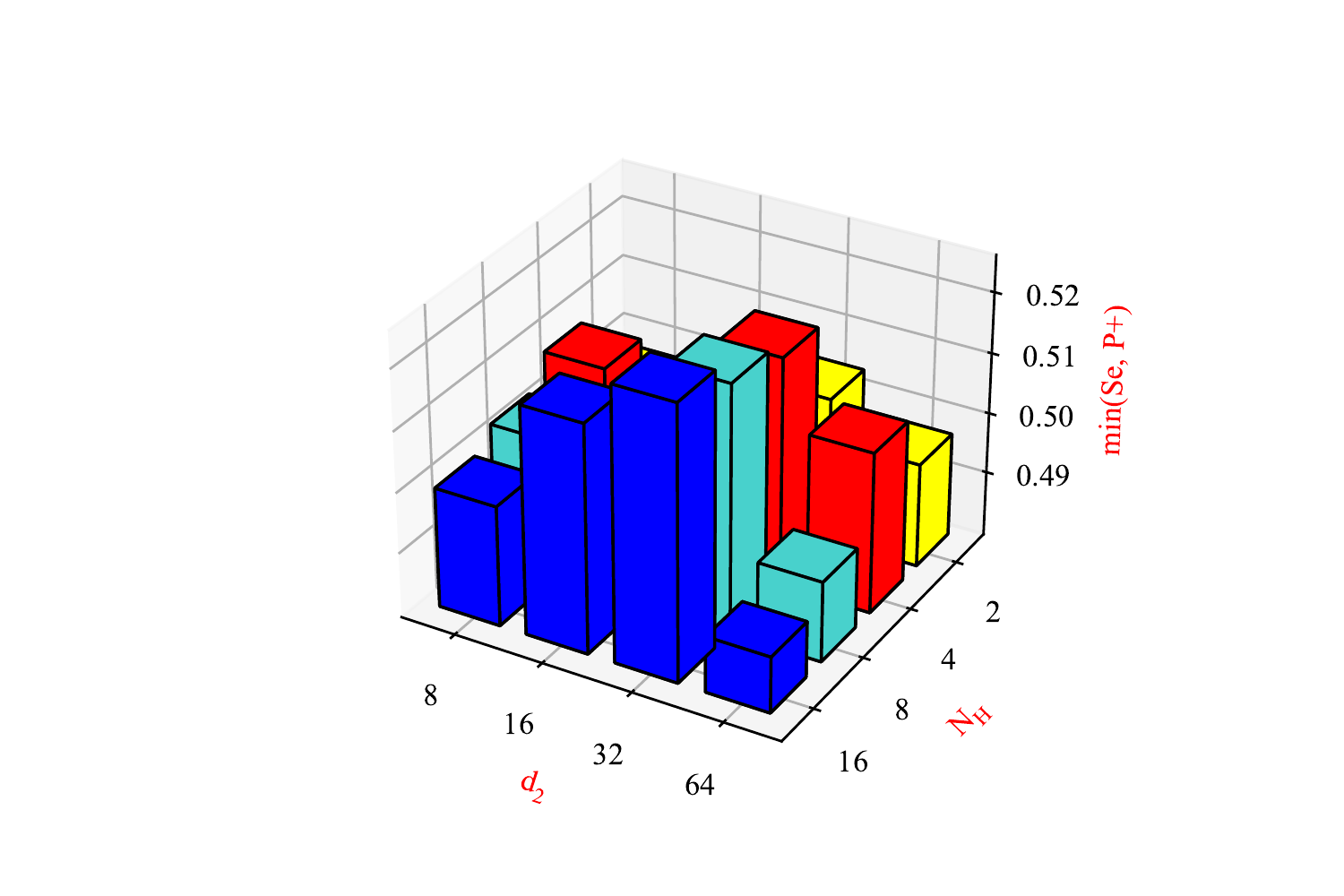} 
	\end{subfigure} 
	\caption{The performance under different hyper-parameter combinations: $d_{2} \times N_{H}$.}
	\label{pse3}
\end{figure*}

\subsection{Experimental Results and Analysis (RQ3)}
\subsubsection{Performance Comparison}
Table \ref{t3} and Fig. \ref{fig3} have shown the experimental results of the HGV and other baselines in two real-world risk prediction tasks. Overall, when compared with all the other methods in performance testing, our proposed HGV consistently achieves the best performance in both tasks. 

Specifically, as we can see, although with good interpretability, both classic machine learning methods and plain RNN-based ones \cite{DBLP:journals/ml/YuHL11,friedman2001greedy} perform poorly. Meanwhile, we also find our model outperforms the time-aware methods \cite{DBLP:conf/nips/ChoiBSKSS16,DBLP:conf/kdd/BaytasXZWJZ17}, which shows that it is not sufficient to only consider the effect of time-aware decay. Furthermore, compared to the HGV, the attention-based baselines \cite{DBLP:conf/emnlp/ChoMGBBSB14,DBLP:conf/icdm/LeePJM18,DBLP:conf/nips/VaswaniSPUJGKP17,DBLP:conf/aaai/SongRTS18,DBLP:conf/aaai/MaZWRWTMGG20} have also shown insufficient performance due to a lack of capturing clip-aware patterns encoded in temporal status correlation. Faced with the challenge of inevitable noise, the performance of the knowledge-enhanced predictor GRASP \cite{DBLP:conf/aaai/ZhangGMWWT21} is still unsatisfactory. 

\begin{figure*}[!h]
	\centering
	\begin{subfigure}{0.35\textwidth}
		\centering 
		\includegraphics[width=\textwidth,height=5cm]{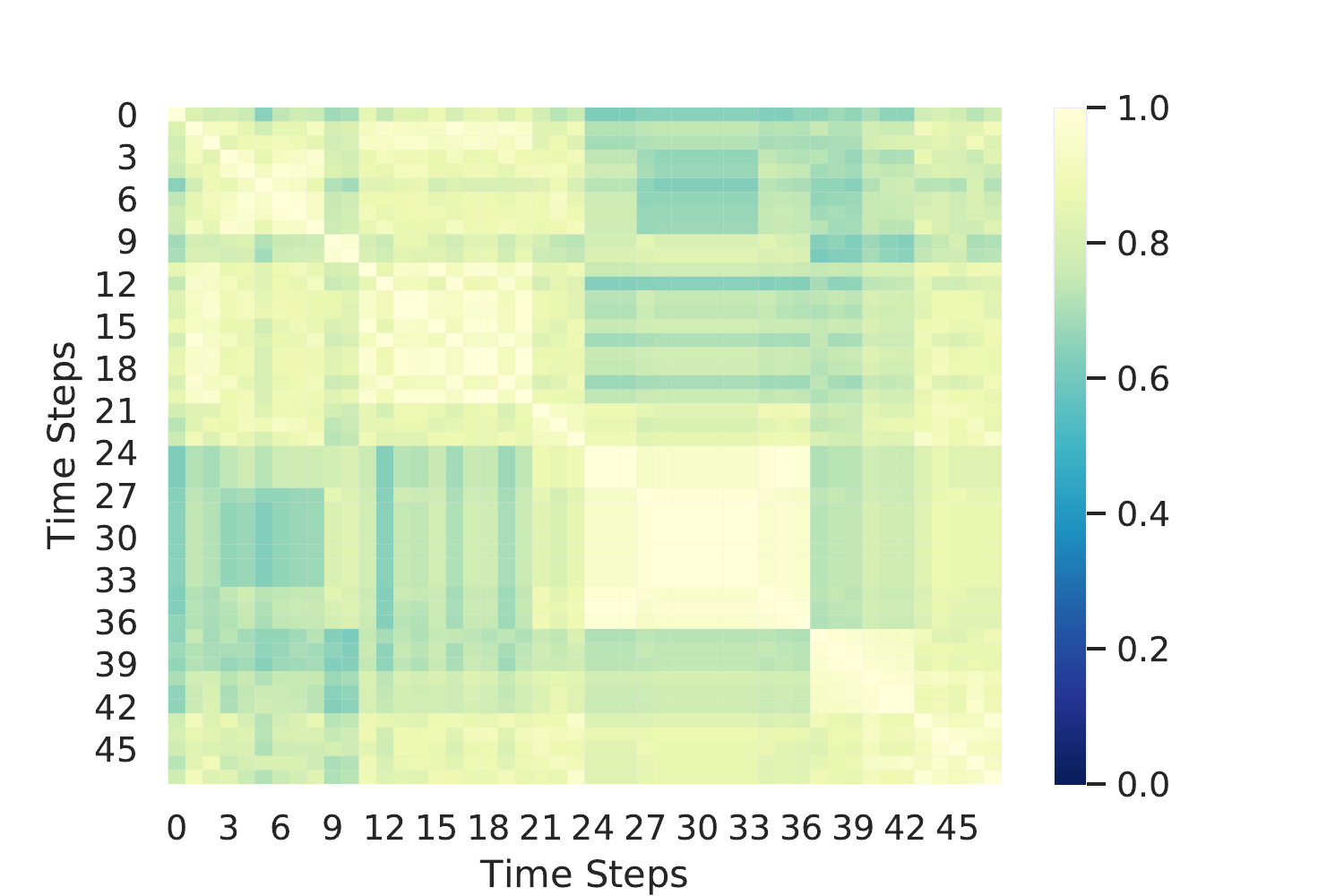} 
	\end{subfigure} 
	\begin{subfigure}{0.3\textwidth} 
		\centering \includegraphics[width=\textwidth,height=5cm]{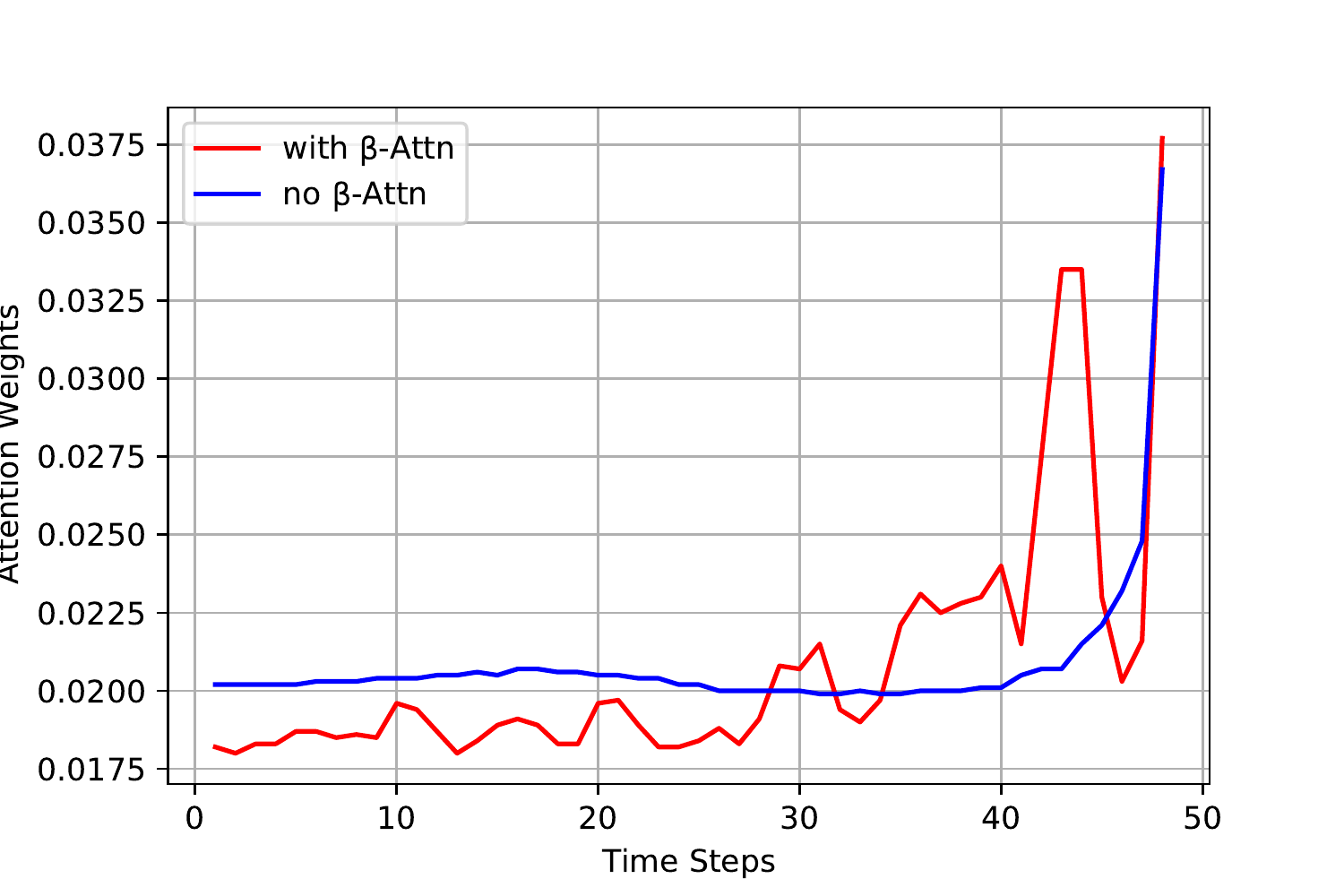} 
	\end{subfigure}
	\caption{Intuitive case analysis for GGE modules and $\beta$-Attn.}
	\label{fig4}
\end{figure*}

\begin{figure*}[ht!]
	\centering
	\includegraphics[width=0.8\linewidth,height=3cm]{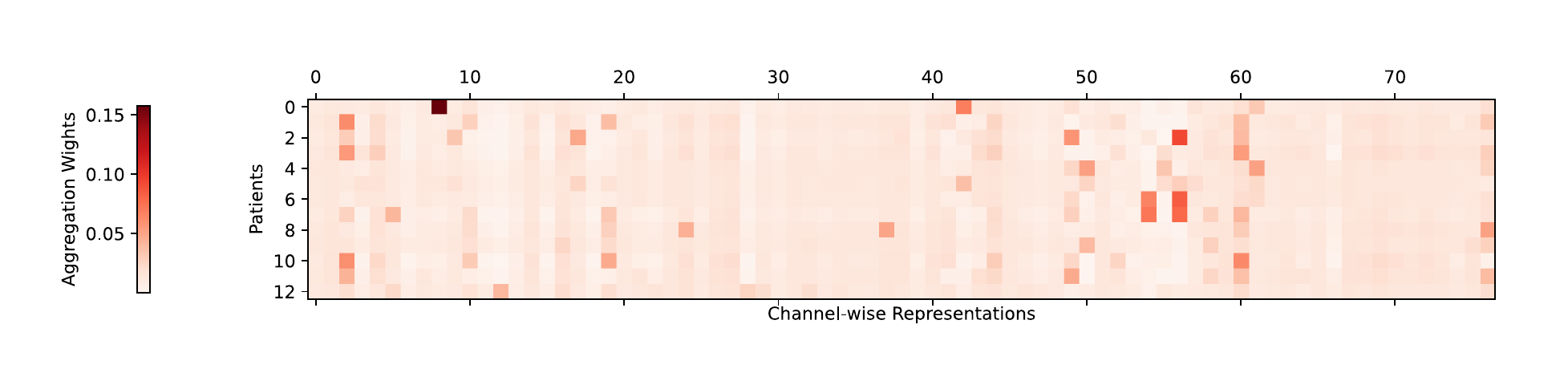}
	\caption{Weights distribution for channel-wise representations in heterogeneous information aggregation.}
	\label{HOTMAP}
\end{figure*}

\begin{figure}[t]
	\centering
	\begin{subfigure}{0.15\textwidth}
		\centering 
		\includegraphics[width=\textwidth,height=2.8cm]{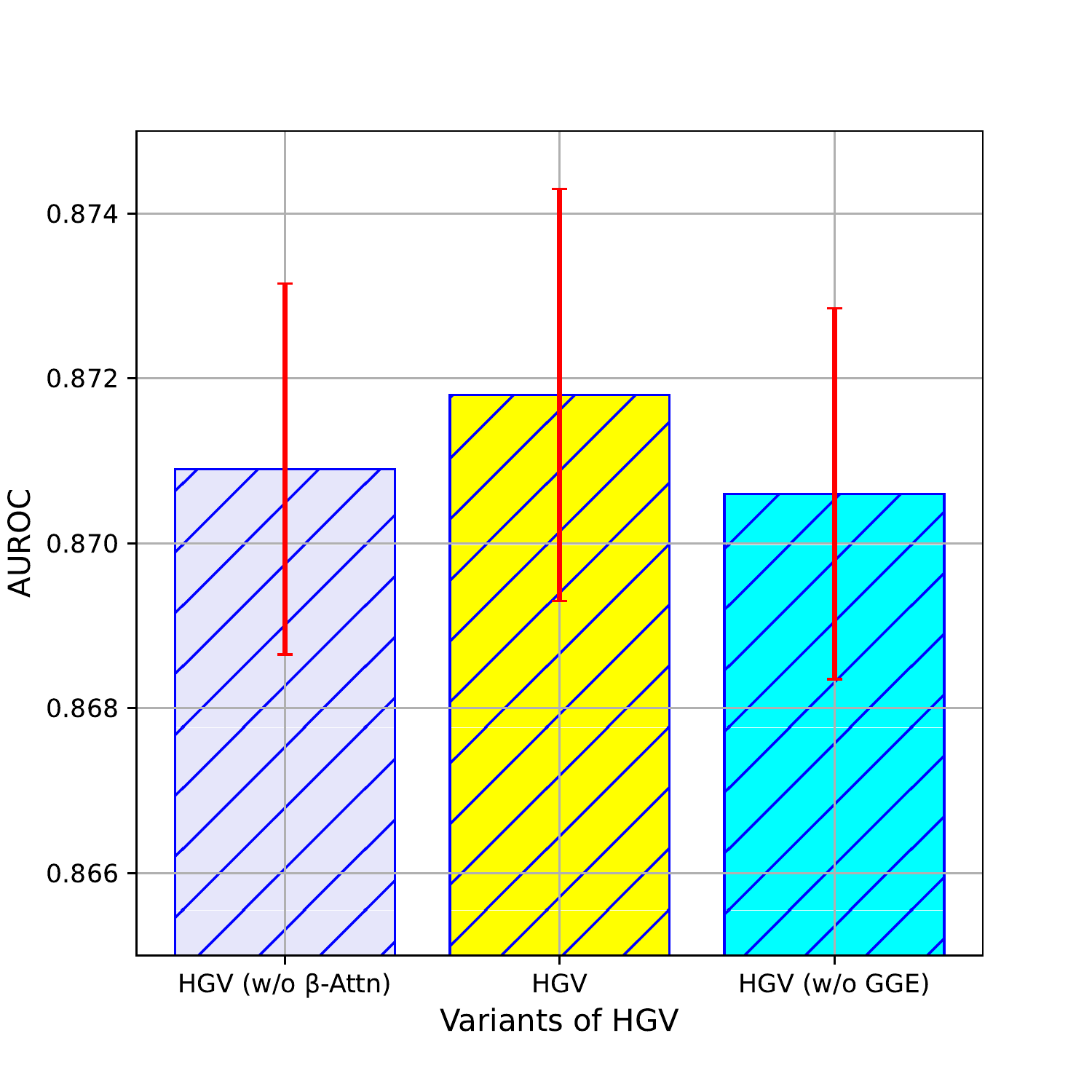} 
	\end{subfigure} 
	\begin{subfigure}{0.15\textwidth} 
		\centering \includegraphics[width=\textwidth,height=2.8cm]{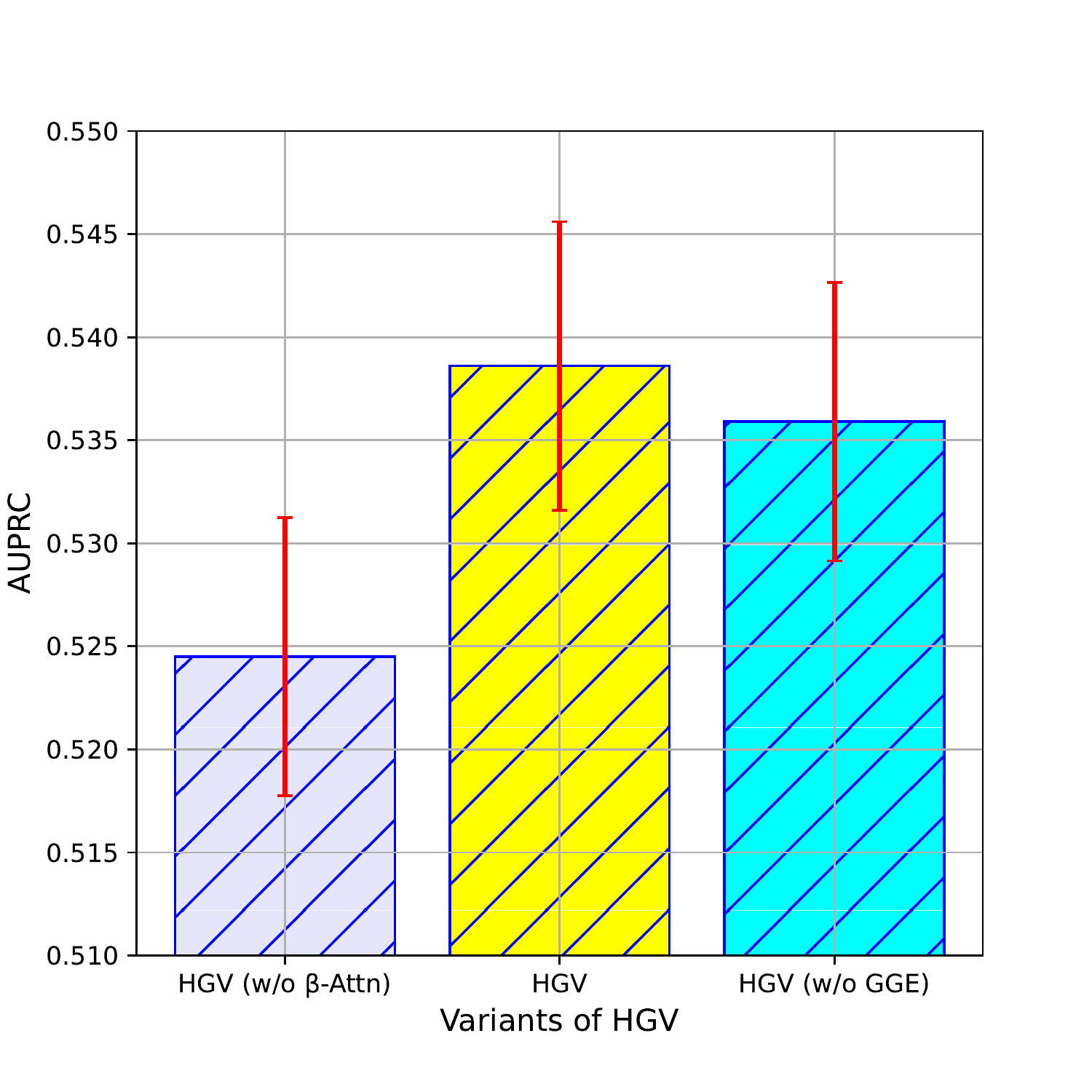} 
	\end{subfigure}
	\begin{subfigure}{0.15\textwidth}
		\centering 
		\includegraphics[width=\textwidth,height=2.8cm]{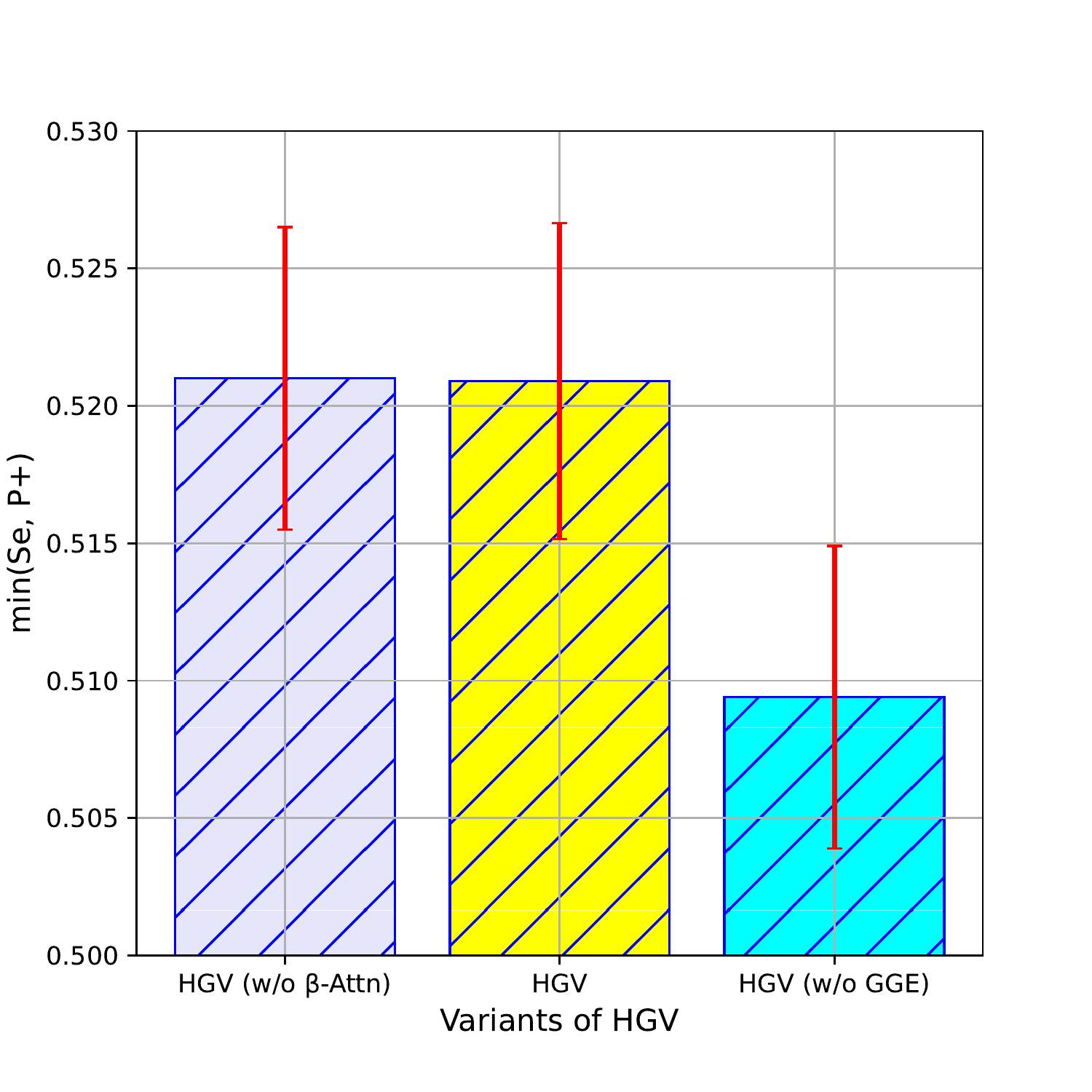} 
	\end{subfigure} 
	\caption{Results of ablation studies.}
	\label{abs}
\end{figure}

Furthermore, to evaluate the performance of the HGV on a more imbalanced and skewed risk prediction task (sparsity is 0.6138 on MIMIC-III but 0.9189 on Ant Group-MYBank), we conduct the financial credit overdue risk prediction experiment on a real-world industrial scenario from MYBank, Ant Group. After analyzing the results on the public benchmark dataset, we select the strongest baseline models as the baselines, namely ConCare \cite{DBLP:conf/aaai/MaZWRWTMGG20} and GRASP \cite{DBLP:conf/aaai/ZhangGMWWT21}. The experimental results in Fig. \ref{fig3} show the AUROC, AUPRC and min(Se, P+) on test set. The proposed model, HGV, outperforms the baselines, and has an average improvement rate of 1.55\%, 0.19\% on AUROC, 1.88\%, 0.92\% on AUPRC and 4.20\%, 2.96\% on min(Se, P+), respectively, compared with ConCare and GRASP. Therefore, we can find that the HGV can still outperform other baselines even in the risk prediction task with more sparsity challenges.

\subsubsection{Parameter sensitivity Analysis}
To show how the main hyper-parameters involved in HGV affect the model performance, we check the sensitivity of some hyper-parameters, the embedding sizes $d_{1}$, $d_{2}$ and the number of head $N_{H}$. Fig. \ref{pse1}-\ref{pse3} show the performance under different hyper-parameter combinations of the proposed HGV model for MIMIC-III dataset. We can see that with a consideration of both efficiency and performance, a relatively smaller number of head (not too small) and larger embedding size (not too large) for the hyper-parameter combinations settings leads to the best result. We take the MIMIC-III dataset as an example, and other datasets also show the similar trend.

\subsubsection{Ablation Studies}
In addition, we also conduct the ablation studies on the benchmark dataset as follows:
\begin{enumerate}
    \item \textbf{HGV (w/o $\beta$-Attn):} To demonstrate the effectiveness of making a global trade-off between time-aware decay and observation significance in sequence representation learning, we replace the $\beta$-Attn with a plain attention module.
    \item \textbf{HGV (w/o GGE):} We also remove the GGE to demonstrate the usefulness of mining heterogeneous correlation beyond homogeneous sequences by constructing the temporal correlation graph.
\end{enumerate}
The results of the ablation study are given in Fig. \ref{abs}, which have proved that both GGE and $\beta$-Attn are effective in the proposed HGV framework.

\subsubsection{Case Study}
To intuitively demonstrate how the main modules, GGE, $\beta$-Attn and heterogeneous information aggregation work in the proposed HGV framework, we give the intuitive case analysis on a randomly selected samples from the MIMIC-III dataset in Fig. \ref{fig4}-\ref{HOTMAP}. 

Specifically, from the left sub-figure in Fig. \ref{fig4}, it is easy to see that there are some local bright blocks in the temporal correlation graph, which shows the clip-aware correlation among different statuses, and figuratively indicates the necessity of extracting such information from time series data. For the second sub-figure, we can also find that the $\beta$-Attn have learned how to make a global trade-off between time-aware decay and observation significance for weighting the time series, which is significantly distinct from the smooth attention distribution learned by only considering time-aware decay.

Furthermore, in Fig. \ref{HOTMAP}, we can find that the heterogeneous information aggregation module can assign different attention weights to each patient for aggregating the channel-wise representations with the hierarchical guidance on two global views.

\section{Conclusion and future work}
This paper proposed a novel end-to-end Hierarchical Global Views-guided sequence representation learning framework (HGV) to predict risk in both healthcare and finance. Specifically, to joint learn hierarchical representations from heterogeneous data, the GGE has achieved to reveal the temporal rhythmic variation of the observed status and the $\beta$-Attn has learned a global trade-off between time-aware decay and observation significance. In addition, we have conducted experiments on two real-world risk prediction tasks and evaluated the performance of the HGV. 

For future work, we will further explore incorporating more explicit prior information into such risk prediction modeling tasks, especially considering the introduction of knowledge graphs into an end-to-end deep learning framework to further improve model interpretability.

\section{Acknowledgments}
% This work was supported in part by Science and Technology Innovation 2030 – "New Generation Artificially Intelligence" Major Project under Grant 2018AAA0102101, in part by the National Natural Science Foundation of China under Grant No. 61976018 and also in part by MyBank, Ant Group. 
This work was supported in part by Science and Technology Innovation 2030 – New Generation Artificial Intelligence Major Project under Grant No. 2018AAA0102100, National Natural Science Foundation of China (Grant No. 61976018, U1936212, 62120106009), and Beijing Natural Science Foundation under Grant No. 7222313. We also thank the support from MyBank, Ant Group.

\newpage

 

% \end{IEEEbiography}



% \vfill

\end{document}